\documentclass[manuscript,screen]{acmart}

\setcopyright{none}
\acmJournal{JACM}

\AtBeginDocument{%
  }

\usepackage{amsmath,amsthm}
\usepackage{mathrsfs}
\usepackage{enumitem}
\usepackage{hyperref}
\usepackage{cleveref}
\usepackage{booktabs}
\usepackage{graphicx}
\usepackage{subcaption}
\usepackage{xcolor}
\usepackage{tikz}
\usepackage{tkz-euclide}
\usepackage{pgfplots}
\pgfplotsset{compat=1.17}
\usetikzlibrary{shapes.geometric, arrows.meta, positioning, shapes,fadings, decorations.pathmorphing, backgrounds, shadows,patterns,calc,3d,bending}
\usepackage{algorithm}
\usepackage{algpseudocode}
\newtheorem{theorem}{Theorem}[section]
\newtheorem{lemma}[theorem]{Lemma}
\newtheorem{proposition}[theorem]{Proposition}
\newtheorem{corollary}[theorem]{Corollary}
\theoremstyle{definition}
\newtheorem{definition}[theorem]{Definition}

\newtheorem{remark}[theorem]{Remark}

\newcommand{\UM}{\textsf{UM}}

\newcommand{\TM}{\textsf{TM}}

\newcommand{\Lib}{\mathscr{L}}
\newcommand{\Width}{\mathcal{W}}
\newcommand{\R}{\mathbb{R}}
\newcommand{\Z}{\mathbb{Z}}
\newcommand{\M}{\mathbb{M}}

\newcommand{\eps}{\varepsilon}

\newcommand{\Hdone}{\mathcal{H}^{d-1}}

\newcommand{\Heven}{\mathcal{H}_{\text{even}}}
\newcommand{\Hodd}{\mathcal{H}_{\text{odd}}}

\begin{document}

\title{The Urysohn Machine: A Metric-Topological Model of Computation}

\author{Xin Li}
\email{xli48@albany.edu}
\orcid{0000-0003-2067-2763}
\affiliation{%
  \institution{University at Albany, State University of New York}
  \city{Albany}
  \state{New York}
  \country{USA}
}

\renewcommand{\shortauthors}{Li}

\begin{abstract}
We introduce the Urysohn Machine~(\UM{}), an effective model of
classification-oriented computation in which metric separation, frontier
structure, and contraction are represented explicitly in the computational
state.  The basic object is a \emph{Urysohn Triple}: a support region, a target
partition, and a separating classifier stored in a reusable Metric Library. 
The topological foundation is a constructive Urysohn Realization theorem for
finite simplicial settings.  It builds a separator from a dyadic ladder of
nested polyhedral regions and equips the resulting frontiers with a chain-level
calculus: each frontier is a cycle, and shells between levels have boundaries
given by differences of frontiers.  This construction leads to two related
complexity measures.  The \emph{decision-boundary width}~\(\Width_{\partial}\)
is the geometric measure of a single classifier's boundary, whereas the
\emph{Urysohn width}~\(\Width_U\) is the total frontier mass represented by a
library or realization.  We prove an Amortized Separation Theorem showing that
approximating a boundary of width~\(\Width_{\partial}\) to accuracy~\(\eps\)
requires a number of simple basis triples proportional to the boundary width and
inversely proportional to the resolution, under explicit boundary-footprint
coverage assumptions.
We further introduce a contrastive separation operator whose graph-cut
functional consistently estimates decision-boundary width from sampled metric
data, while its Laplacian spectrum certifies class-component structure and
conductance.  Finally, we analyze the dynamic Urysohn ladder and establish four
computational guarantees: separability under quotient collapse, stability of
committed frontiers, bounded capacity under contraction, and scalability with
quotient distance.  Together, these results provide a metric-topological account
of classification complexity, amortized inference, and compositional reuse that
preserves classical computability while exposing geometric structure hidden by
purely symbolic descriptions.
\end{abstract}

\ccsdesc[500]{Theory of computation~Abstract machines}
\ccsdesc[500]{Mathematics of computing~Geometric topology}

\keywords{cognitive computation, metric topology, Urysohn machine,
  Urysohn width/ladder, decision boundary,
  amortized separation, spectral estimation}

\maketitle

\section{Introduction}\label{sec:intro}

Classical models of computation, including Turing machines~(\TM{}s)
\cite{turing1936computable}, the $\lambda$-calculus~\cite{church1936unsolvable},
and recursive-function models~\cite{kleene1936general}, describe computation in
terms of symbolic states and local rewrite rules.  These models give a precise
account of computability and support the standard complexity-theoretic hierarchy.
They are intentionally substrate-neutral: geometry, continuity, and distance are
not primitive notions.  For classification over a metric or topological space,
this means that all geometric structure must be encoded indirectly by a program.
Such encodings are universal, but they conflate two different forms of
difficulty: the extrinsic cost of the program that implements a classifier and
the intrinsic cost of the decision boundary that the classifier must resolve.

This paper develops a complementary model, the \emph{Urysohn Machine}~(\UM{}),
for classification-oriented computation in metric spaces.  
The basic object of the model is a \emph{Urysohn Triple}
\((\Sigma,\Pi,f)\), consisting of a support region~\(\Sigma\), a partition of
target concepts~\(\Pi\), and a classifier~\(f\) that separates the parts of the
partition.  A finite stack of such triples forms a \emph{Metric Library}, which
can be queried, refined, frozen, and reused.  The topological motivation comes
from Urysohn's lemma~\cite{urysohn_zum_1925}: in a normal topological space, two
disjoint closed sets admit a continuous separator.  We strengthen this existence
statement into a constructive realization theorem for finite simplicial settings.
The resulting separator is assembled from a dyadic ladder of nested polyhedral
regions, and its frontiers carry a chain-level calculus: each frontier is a
cycle, and shells between dyadic levels have boundaries equal to differences of
frontiers.

Urysohn realization theorem leads to the paper's central complexity measure.  We
distinguish the \emph{decision-boundary width}
\(\Width_{\partial}\), the geometric measure of a single classifier's boundary,
from the \emph{Urysohn width} \(\Width_U\), the total frontier mass represented
by a Urysohn library or realization.  The former measures the geometric size of a
target separator; the latter measures the amount of separation structure that a
particular~\UM{} stores, composes, or reuses.  This distinction captures a form
of intrinsic classification complexity that is invisible to standard Turing
complexity: two classifiers may lie in the same symbolic complexity class while
requiring very different amounts of boundary structure to realize.

Connectivity is the structural invariant behind this construction
\cite{nakahara_geometry_2018}.  The dyadic ladder realizes separation not by
arbitrarily cutting the space, but by inserting graded frontiers that preserve
the relevant connected structure between levels.  Operationally, the same
principle appears in the contrastive separation operator of
\Cref{def:cs-operator}: diffusion is allowed within class-consistent regions but
suppressed across committed boundaries.  Its spectrum therefore certifies
topological information such as the number of class-connected components, while
the associated graph-cut functional estimates boundary measure.  In this sense,
topology decides whether separation is possible \cite{cover1965geometry} and how many connected regions
must be distinguished; the metric prices how much frontier must be laid down to
perform that separation.

The main results bracket the model from two sides.  The Urysohn Realization
Theorem (\Cref{thm:realization}) gives the constructive upper side: separability
can be compiled into a finite, composable, chain-level frontier representation.
The Amortized Separation Theorem (\Cref{thm:amortized}) gives the lower side:
approximating a boundary of width \(\Width_{\partial}\) to accuracy~\(\eps\)
requires a number of simple basis triples proportional to that width and
inversely proportional to the resolution.  Thus a difficult separator may be
expensive to build, but once represented in the Metric Library, future queries
can exploit amortized lookup, local contraction, and quotient-level inference.
The paper then studies the dynamic Urysohn ladder induced by repeated
refinement.  This ladder supports four guarantees: separability, because quotient
collapse preserves class separation; stability, because committed frontiers are
invariant under later refinement; bounded capacity, because covering numbers per
quotient level remain controlled under contraction; and scalability, because
inference cost scales with quotient distance rather than ambient distance.  These
guarantees explain what the~\UM{} adds over an ordinary symbolic implementation:
not more computability, but a structured representation in which metric
separation, topological invariants, and amortized inference are explicit.
In summary, we make the following contributions in this paper.
\begin{enumerate}[leftmargin=*,nosep]
  \item We define the Urysohn Machine, including Urysohn Triples, Metric
  Libraries, stack operations, and the associated notion of cognitive
  computation.
  \item We prove a constructive Urysohn Realization theorem that exposes the
  dyadic and chain-level structure of continuous separation.
  \item We distinguish decision-boundary width from Urysohn width and prove
  an Amortized Separation Theorem relating boundary measure to library size.
  \item We introduce the contrastive separation operator, whose graph-cut
  functional estimates boundary measure and whose spectrum certifies
  class-component structure.
  \item We develop the dynamic Urysohn ladder view of computation and derives
  separability, stability, bounded-capacity, and scalability guarantees.
  \item We establish four computational guarantees, including separability, stability, bounded capacity, and scalability. Together, they become the theoretical foundation for cognitive computation using ~\UM{}.
\end{enumerate}



\section{The Urysohn Machine}\label{sec:um}

\paragraph{The Metric Library}

The~\UM{}'s computational substrate is the Metric Library, a structured space that serves
simultaneously as memory, program, and workspace.

\begin{definition}[Metric Library]\label{def:library}
  A \emph{Metric Library} is a 5-tuple $\Lib = (S,d,T,\sigma,K)$ where:
  \begin{enumerate}
    \item $S$ is a countable discrete space---the \emph{base concepts space};
    \item $d\colon S \times S \to \R_{\geq 0}$ is a metric guaranteeing that Urysohn's
      lemma applies;
    \item $T$ is a finite collection of Urysohn Triples (defined below);
    \item $\sigma$ is a stack discipline governing triple access; and
    \item $K \in \Z^{+}$ is a fixed constant bounding $|T|$.
  \end{enumerate}
Elements of~$S$ are \emph{indices} - discrete labels for topological regions. The
  library is fixed at initialization; the operator cannot resize it.
\end{definition}

The requirement that $S$ be discrete with metric~$d$ is essential. It guarantees that $S$
is metrizable (indeed, that Urysohn's lemma~\cite{urysohn_zum_1925} applies in the relevant
sense) while keeping the base structure countable. Each element of~$S$ carries both
identity (as an index) and relational structure (through~$d$). This dual
role (combinatorial and geometric) is what distinguishes the Metric Library from a
\TM{}'s tape.

\paragraph{Urysohn Triples} The name ``Urysohn Triple'' reflects the connection to Urysohn's lemma \cite{urysohn_zum_1925}. In a normal
topological space, any two disjoint closed sets can be separated by a continuous function.
Each triple instantiates this principle: it provides a concrete separating function for a
specific pair of closed regions within its support. The~\UM{}'s computational power
derives from the ability to compose, stack, and sequence these local separators. 

\begin{definition}[Urysohn Triple]\label{def:triple}
  A \emph{Urysohn Triple} is a 3-tuple $\tau = (\Sigma, \Pi, f)$ where:
  \begin{enumerate}
    \item $\Sigma \subseteq S$ is the \emph{support}, the region of~$S$ where the triple
      is defined;
    \item $\Pi = \{P_1, P_2, \ldots, P_k\}$ is the \emph{target partition}, the set of
      concepts this triple discriminates; and
    \item $f\colon \Sigma \to \Pi$ is the \emph{classifier}, a function mapping points
      in the support to partition labels.
  \end{enumerate}
  Each triple is a perfect separator for its target partition: for every pair of labels
  $P_i \neq P_j$, the triple correctly classifies every point in its support into the
  corresponding region.
\end{definition}

\begin{figure}[h]
\centering
\begin{tikzpicture}[
  triple/.style={
    draw,
    rounded corners=3pt,
    minimum width=3.3cm,
    minimum height=0.72cm,
    fill=#1!15,
    font=\small
  },
  arr/.style={-{Stealth[length=5pt]}, thick},
  op/.style={
    font=\footnotesize\bfseries,
    rounded corners=2pt,
    inner sep=2pt
  },
  note/.style={
    font=\scriptsize,
    align=center,
    text=gray!75!black
  }
]

\node[triple=blue] (t1) at (0,0) {$\tau_1=(\Sigma_1,\Pi_1,f_1)$};
\node[triple=green!70!black] (t2) at (0,0.95) {$\tau_2=(\Sigma_2,\Pi_2,f_2)$};
\node[triple=orange] (t3) at (0,1.90) {$\tau_3=(\Sigma_3,\Pi_3,f_3)$};

\draw[thick, gray] (-1.90,-0.45) -- (-1.90,2.42);
\draw[thick, gray] ( 1.90,-0.45) -- ( 1.90,2.42);
\draw[thick, gray] (-1.90,-0.45) -- ( 1.90,-0.45);

\node[font=\scriptsize\bfseries, text=orange!80!black, anchor=west]
  at (2.12,1.80) {top / active};

\node[font=\scriptsize\itshape, text=gray!70!black, anchor=west]
  at (2.12,0.48) {frozen below top};

\draw[arr] (-4.20,1.90) -- node[above, font=\footnotesize] {$x$} (-1.50,1.95);
\draw[arr] (1.90,1.90) -- node[above, font=\footnotesize] {$\hat y=f_3(x)$} (3.80,1.95);

\node[op, fill=green!12, text=green!45!black]
  at (0,2.68) {\textsc{Read}};
\node[note] at (0,3.05)
  {evaluate using the\\top triple only};

\node[triple=purple] (new) at (-3.45,3.25)
  {$\tau_4=(\Sigma_4,\Pi_4,f_4)$};

\node[op, fill=blue!12, text=blue!70!black]
  at (-3.75,2.15) {\textsc{Push}};

\draw[arr, blue!70!black]
  (new.south) -- node[right, font=\scriptsize, text=blue!70!black]
  {onto top} (-3.45,2.02) -- (-1.90,2.02);

\node[note, text=blue!70!black] at (-3.45,3.88)
  {new context\\becomes active};

\node[op, fill=red!12, text=red!70!black]
  at (3.75,2.15) {\textsc{Pop}};

\draw[arr, red!70!black]
  (1.90,2.02) -- (3.45,2.02) -- node[right, font=\scriptsize, text=red!70!black]
  {remove top} (3.45,3.05);

\node[note, text=red!70!black] at (3.45,3.45)
  {revert to\\$\tau_2$};

\node[font=\scriptsize, text=gray!60!black] at (0,-0.75)
  {bottom of stack};

\end{tikzpicture}
\caption{Stack architecture of the \UM{}. Urysohn triples are stored in
last-in-first-out order. \textsc{Push} inserts a new triple onto the top of
the stack, making it the active context. \textsc{Pop} removes the current
top triple and restores the previous context. \textsc{Read} is non-mutating:
the input \(x\) is classified only by the current top triple, here
\(\tau_3\), while lower triples remain frozen.}
\Description{Stack diagram showing Urysohn triples stored in last-in-first-out order. Push inserts a new triple from above, pop removes the top triple upward, and read evaluates the input using only the top triple.}
\label{fig:stack}
\end{figure}

\paragraph{Stack architecture}

The stack discipline enforces a structural constraint (Fig. \ref{fig:stack}): contexts are nested, not
interleaved. When a new classification context arises, the operator pushes a fresh triple;
when that context ends, popping restores the previous classifier exactly. This models the
natural structure of hierarchical classification tasks \cite{simon1973architecture}: coarse-grained decisions are made
first, and fine-grained refinements are layered on top, then removed when no longer needed.
Note that the choice of a stack rather than a queue is dictated by the topology of the
construction. The supports of the triples are \emph{nested}: by the dyadic
construction of \Cref{thm:realization}, each refinement inserts a separating
level whose support is contained in that of every coarser level already present,
$\overline{U(r)}\subset U(s)$ for $r<s$. Containment is a partial order, and
resolving the innermost (finest) region before returning to the enclosing
(coarser) one is precisely a last-in-first-out discipline: \textsc{Push} enters a
finer classifier valid only \emph{within} the current support, and \textsc{Pop}
restores the enclosing classifier unchanged. Because past triples are frozen
(Definition \ref{def:operator}), the stack is a scope mechanism in the sense of a call
stack: a committed coarse decision is the environment inside which all finer
decisions live. 
Beyond this
nesting argument, the stack is also the space-economical choice - i.e., its depth, rather than the width of the frontier is the resource consumed, a point we will make
precise in \Cref{rem:stack-savitch}.


\begin{definition}[\UM{} Operator]\label{def:operator}
  The \emph{\UM{} Operator} is a deterministic function $O\colon \Lib \to \Lib$ that may:
  \begin{enumerate}
    \item \textsc{Push} a new triple~$\tau$ onto the stack;
    \item \textsc{Pop} the most recent triple, reverting to the previous one; or
    \item \textsc{Read} the top-of-stack triple for classification.
  \end{enumerate}
  The operator uses only the top-of-stack triple for classification. \textsc{Pop} removes
  the most recent triple, reverting to the previous one. Past triples are
  frozen; they cannot be modified.
\end{definition}


\begin{definition}[\UM{} Computation]\label{def:computation}
  A \emph{\UM{} computation} is a sequence
  $C = (\Lib_0, O_1, \Lib_1, \ldots)$ where $\Lib_0$ is the initial
  library state, each $O_i$ is an operator application, and
  $\Lib_i = O_i(\Lib_{i-1})$. The computation \emph{converges} if there exists~$N$ such
  that for all $n > N$, the top-of-stack classification is identical. It \emph{diverges}
  otherwise.
\end{definition}


\section{Topological Foundation: The Fundamental Theorem of Urysohn Realization}\label{sec:foundation}

The~\UM{}'s claim that classification reduces to the construction of continuous separating
functions rests on Urysohn's lemma. But the classical lemma is a pure existence result:
it guarantees that a separator $f: X \to [0,1]$ exists for any two disjoint closed sets
in a normal space, without specifying the algebraic structure of $f$'s level sets. For the~\UM{} to operate constructively, building separators from data, composing them in a
library, measuring their complexity via Urysohn Width, we need a \emph{constructive}
version that makes the chain-level structure of the separator explicit. The Fundamental
Theorem of Urysohn Realization provides exactly this.

\paragraph{Dyadic normality and frontier realization}

The constructive Urysohn proof builds the separator by dyadic refinement: nested open
sets indexed by dyadic rationals $r \in [0,1] \cap \mathbb{D}$, with successive levels
inserted between previously constructed levels. We decompose this construction into three
independent axioms.

\begin{definition}[Dyadic normality]\label{def:dyadic-normality}
Let $K$ be a finite simplicial complex with $X = |K|$, and let $A, B \subset X$ be
disjoint closed subpolyhedra. An admissible class $\mathcal{R}$ of regular closed
polyhedral neighborhoods is \emph{dyadically normal relative to $(A,B)$} if it
satisfies:
\begin{itemize}[nosep]
  \item \textbf{T1} (Initial separation): $\exists P_0 \in \mathcal{R}$ with
    $A \subset \mathrm{Int}_X P_0$ and $P_0 \cap B = \varnothing$.
  \item \textbf{T2} (Binary interpolation): $P \Subset Q \implies \exists R \in
    \mathcal{R}$ with $P \Subset R \Subset Q$.
  \item \textbf{T3} (Terminal interpolation): $P \cap B = \varnothing \implies \exists
    Q \in \mathcal{R}$ with $P \Subset Q$ and $Q \cap B = \varnothing$.
\end{itemize}
\end{definition}

Axioms T1--T3 are the constructive content of classical normality. They supply the
topological room to build nested dyadic neighborhoods. The second ingredient connects
these neighborhoods to the chain complex. We write
$\mathrm{Fr}_X(P) := \overline{P} \cap \overline{X \setminus P}$ for the
\emph{topological frontier} of~$P$ in~$X$ (the set of points adherent to both $P$ and
its complement).

\begin{definition}[Frontier realization]\label{def:frontier-realization}
The admissible class $\mathcal{R}$ admits \emph{chain-level frontier realization} if every
$P \in \mathcal{R}$ carries a chain $c_P \in C_d(K_P; \mathbf{k})$ on some subdivision
$K_P$ of $K$ such that: 1) the algebraic boundary $\partial c_P$ realizes the
topological frontier, $|\partial c_P| = \mathrm{Fr}_X(P)$; and 2) for $P \Subset Q$,
the shell chain $s_{P,Q} := c_Q - c_P$ has support
$|s_{P,Q}| = \overline{Q \setminus \mathrm{Int}_X P}$.
\end{definition}

The third ingredient is the algebraic coherence identity $\partial^2 = 0$ \cite{wheeler1990information}, the defining
property of a chain complex \cite{edelsbrunner_computational_2010}. In the simplicial setting, it is automatic; we state it
explicitly to isolate its precise role.

\paragraph{The Fundamental Theorem}
Before stating the main result, we first clarify the role of the theorem in the
overall argument.  Classical Urysohn separation asserts that, in a normal
topological space, two disjoint closed sets can be separated by a continuous
function into the unit interval.  The result below strengthens this statement in
the finite simplicial setting by realizing the separator through a nested dyadic family of polyhedral regions
whose frontiers carry an explicit chain-level algebra.
As shown in Fig. \ref{fig:dyadic}, the dyadic
sets \(P_r\) provide the topological refinement needed to construct the separator
\(f\), while their simplicial frontiers \(\Gamma_r=\partial c_{P_r}\) provide the
homological bookkeeping needed to track how separation changes across levels.
Because \(\partial^2=0\) \cite{wheeler1990information}, each frontier is automatically a cycle, and the shell
between two dyadic levels has boundary equal to the difference of the two
frontiers.  The theorem converts Urysohn separation into a coherent
``frontier calculus'': separation is represented simultaneously as a continuous
function, a nested polyhedral filtration, and a compatible family of cycles and
shell chains.

\begin{theorem}[Fundamental Theorem of Urysohn Realization]\label{thm:realization}
Let $K$ be a finite simplicial complex with $X = |K|$, let $A, B \subset X$ be disjoint
closed subpolyhedra, and let $\partial$ denote the simplicial boundary operator with
$\partial^2 = 0$. Suppose $(X, \mathcal{R})$ is dyadically normal relative to $(A,B)$
and $\mathcal{R}$ admits chain-level frontier realization. Then:
1) There exists a continuous $f: X \to [0,1]$ with $f|_A = 0$ and $f|_B = 1$,
    constructed from a dyadic family $\{P_r\}_{r \in \mathbb{D} \cap [0,1)}$ via
    $f(x) = \inf\{r : x \in \mathrm{Int}_X P_r\}$;
2) Every dyadic frontier $\Gamma_r := \partial c_{P_r}$ is a cycle:
    $\partial \Gamma_r = 0$;
3) For $r < s$, the shell chain $s_{r,s} = c_{P_s} - c_{P_r}$ satisfies
    $\partial s_{r,s} = \Gamma_s - \Gamma_r$ and $\partial(\Gamma_s - \Gamma_r) = 0$.
The dyadic separator $f$ is accompanied by an algebraically coherent chain-level
frontier calculus.
\end{theorem}

\begin{figure}[ht]
\centering
\begin{subfigure}[b]{0.42\textwidth}
  \centering
  \resizebox{\linewidth}{!}{%
\begin{tikzpicture}[scale=0.95]
\fill[blue!5] (0,0) ellipse (4.5 and 2.0);
\fill[blue!10] (0,0) ellipse (4.2 and 1.85);
\fill[blue!15] (0,0) ellipse (3.6 and 1.6);
\fill[blue!22] (0,0) ellipse (2.5 and 1.15);
\fill[blue!30] (0,0) ellipse (1.4 and 0.65);
\fill[blue!40] (0,0) ellipse (0.8 and 0.4);
 
\fill[blue!60] (-0.15,0) ellipse (0.3 and 0.2);
\node[font=\small\bfseries, white] at (-0.15,0) {$A$};
 
\fill[red!40] (5.0,0) ellipse (0.45 and 0.3);
\node[font=\small\bfseries] at (5.0,0) {$B$};
 
\node[font=\scriptsize, anchor=south west] at (0.30, 0.15) {$P_0$};
\node[font=\scriptsize, anchor=south west] at (0.50, 0.36) {$P_{1/4}$};
\node[font=\scriptsize, anchor=south west] at (1.10, 0.75) {$P_{1/2}$};
\node[font=\scriptsize, anchor=south west] at (1.60, 1.10) {$P_{3/4}$};
\node[font=\scriptsize, anchor=south west] at (2.25, 1.35) {$P_{7/8}$};
 
\draw[blue!60, thick] (0,0) ellipse (0.8 and 0.4);
\draw[blue!50, thick, dashed] (0,0) ellipse (1.4 and 0.65);
\draw[blue!40, thick] (0,0) ellipse (2.5 and 1.15);
\draw[blue!30, thick, dashed] (0,0) ellipse (3.6 and 1.6);
\draw[blue!20, thick] (0,0) ellipse (4.2 and 1.85);
 
\draw[-{Stealth[length=4pt]}, thick, gray] (-0.5,-2.5) -- (5.5,-2.5);
\node[font=\scriptsize, gray] at (-0.5,-2.8) {$f=0$};
\node[font=\scriptsize, gray] at (5.0,-2.8) {$f=1$};
\foreach \x/\lab in {0.8/0, 1.4/\frac{1}{4}, 2.5/\frac{1}{2},
                      3.6/\frac{3}{4}, 4.2/\frac{7}{8}} {
  \draw[gray] (\x,-2.35) -- (\x,-2.65);
  \fill[gray] (\x,-2.5) circle (1.2pt);
  \node[font=\tiny, gray, below] at (\x,-2.65) {$\lab$};
}
\node[font=\scriptsize, blue!60!black] at (-2.6, -1.0) {$\Gamma_r = \partial c_{P_r}$};
\draw[-{Stealth[length=3pt]}, blue!60!black, thin] (-1.6,-0.9) -- (-0.9,-0.4);
\end{tikzpicture}}
  \caption{Dyadic nesting in the Urysohn construction. The neighborhoods
$\{P_r\}_{r \in \mathbb{D} \cap [0,1)}$ nest compactly around the closed set~$A$
and remain disjoint from~$B$. Each frontier $\Gamma_r = \partial c_{P_r}$ (solid and
dashed ellipses) is a cycle by algebraic coherence ($\partial^2 = 0$). The separator
$f(x) = \inf\{r : x \in \mathrm{Int}\, P_r\}$ maps $A$ to $0$ and $B$ to $1$, with
level sets determined by the nesting structure. Note that $P_{3/4}$ and $P_{7/8}$ are
closer together than earlier levels, reflecting the dyadic refinement in which each step
bisects the gap between existing levels.}
  \label{fig:dyadic}
\end{subfigure}
\hfill
\begin{subfigure}[b]{0.56\textwidth}
  \centering
  \resizebox{\linewidth}{!}{%
\begin{tikzpicture}[
  rung/.style={draw, thick, fill=#1!25, minimum width=2.0cm, minimum height=0.22cm,
    inner sep=0pt},
  lbl/.style={font=\scriptsize, anchor=west},
  phase/.style={font=\scriptsize\itshape, text=#1},
  arr/.style={-{Stealth[length=5pt]}, thick, gray!60}
]
\begin{scope}[shift={(-5,0)}]
  \node[font=\footnotesize\bfseries] at (1,4.3) {Level 0};
  \fill[blue!50] (0,0) rectangle (2,0.3);
  \node[font=\scriptsize, white] at (1,0.15) {$A$};
  \fill[red!40] (0,3.7) rectangle (2,4.0);
  \node[font=\scriptsize] at (1,3.85) {$B$};
  \node[rung=green!60!black] at (1,2.0) {};
  \node[lbl] at (2.2,2.0) {$U(\tfrac{1}{2})$};
  \draw[gray!40, thick] (0,0.3) -- (0,3.7);
  \draw[gray!40, thick] (2,0.3) -- (2,3.7);
  \node[font=\tiny, gray] at (-0.3,0.15) {$0$};
  \node[font=\tiny, gray] at (-0.3,2.0) {$\tfrac{1}{2}$};
  \node[font=\tiny, gray] at (-0.3,3.85) {$1$};
\end{scope}
 
\draw[arr] (-2.5,2.0) -- (-1.5,2.0);
\node[phase=orange!80!black] at (-2.0,2.5) {Detect};
\node[phase=blue!70!black] at (-2.0,1.5) {Refine};
 
\begin{scope}[shift={(0,0)}]
  \node[font=\footnotesize\bfseries] at (1,4.3) {Level 1};
  \fill[blue!50] (0,0) rectangle (2,0.3);
  \node[font=\scriptsize, white] at (1,0.15) {$A$};
  \fill[red!40] (0,3.7) rectangle (2,4.0);
  \node[font=\scriptsize] at (1,3.85) {$B$};
  \node[rung=green!70!black] at (1,1.0) {};
  \node[lbl] at (2.2,1.0) {$U(\tfrac{1}{4})$};
  \node[rung=green!60!black] at (1,2.0) {};
  \node[lbl] at (2.2,2.0) {$U(\tfrac{1}{2})$};
  \node[rung=green!50!black] at (1,3.0) {};
  \node[lbl] at (2.2,3.0) {$U(\tfrac{3}{4})$};
  \draw[gray!40, thick] (0,0.3) -- (0,3.7);
  \draw[gray!40, thick] (2,0.3) -- (2,3.7);
  \node[font=\tiny, gray] at (-0.3,0.15) {$0$};
  \node[font=\tiny, gray] at (-0.3,1.0) {$\tfrac{1}{4}$};
  \node[font=\tiny, gray] at (-0.3,2.0) {$\tfrac{1}{2}$};
  \node[font=\tiny, gray] at (-0.3,3.0) {$\tfrac{3}{4}$};
  \node[font=\tiny, gray] at (-0.3,3.85) {$1$};
  \draw[orange, thick, dashed] (-0.1,0.88) rectangle (2.1,1.12);
  \draw[orange, thick, dashed] (-0.1,2.88) rectangle (2.1,3.12);
\end{scope}
 
\draw[arr] (2.5,2.0) -- (3.5,2.0);
\node[phase=orange!80!black] at (3.0,2.5) {Detect};
\node[phase=blue!70!black] at (3.0,1.5) {Refine};
 
\begin{scope}[shift={(5,0)}]
  \node[font=\footnotesize\bfseries] at (1,4.3) {Level 2};
  \fill[blue!50] (0,0) rectangle (2,0.3);
  \node[font=\scriptsize, white] at (1,0.15) {$A$};
  \fill[red!40] (0,3.7) rectangle (2,4.0);
  \node[font=\scriptsize] at (1,3.85) {$B$};
  \node[rung=green!80!black] at (1,0.5) {};
  \node[lbl] at (2.2,0.5) {$\tfrac{1}{8}$};
  \node[rung=green!70!black] at (1,1.0) {};
  \node[lbl] at (2.2,1.0) {$\tfrac{1}{4}$};
  \node[rung=green!65!black] at (1,1.5) {};
  \node[lbl] at (2.2,1.5) {$\tfrac{3}{8}$};
  \node[rung=green!60!black] at (1,2.0) {};
  \node[lbl] at (2.2,2.0) {$\tfrac{1}{2}$};
  \node[rung=green!55!black] at (1,2.5) {};
  \node[lbl] at (2.2,2.5) {$\tfrac{5}{8}$};
  \node[rung=green!50!black] at (1,3.0) {};
  \node[lbl] at (2.2,3.0) {$\tfrac{3}{4}$};
  \node[rung=green!45!black] at (1,3.4) {};
  \node[lbl] at (2.2,3.4) {$\tfrac{7}{8}$};
  \draw[gray!40, thick] (0,0.3) -- (0,3.7);
  \draw[gray!40, thick] (2,0.3) -- (2,3.7);
  \node[font=\tiny, gray] at (-0.3,0.15) {$0$};
  \node[font=\tiny, gray] at (-0.3,3.85) {$1$};
  \draw[orange, thick, dashed] (-0.1,0.38) rectangle (2.1,0.62);
  \draw[orange, thick, dashed] (-0.1,1.38) rectangle (2.1,1.62);
  \draw[orange, thick, dashed] (-0.1,2.38) rectangle (2.1,2.62);
  \draw[orange, thick, dashed] (-0.1,3.28) rectangle (2.1,3.52);
\end{scope}
 
\node[font=\scriptsize, align=center, text=gray!70!black] at (1,-0.7)
  {Refinement round: \textbf{Evaluate} (classify) $\to$
   \textbf{Detect} (find coarse gap) $\to$ \textbf{Refine} (insert new rungs,
   freeze existing)};
\node[font=\scriptsize, align=center, text=green!50!black] at (1,-1.3)
  {At convergence: rungs dense in $[0,1]$, separator $f$ continuous,
   no further refinement triggered};
\end{tikzpicture}}
  \caption{The Urysohn Ladder as incremental separator construction. Each level of dyadic refinement
corresponds to one round of Evaluate-Detect-Refine. Level~0 inserts one rung
$U(1/2)$; Level~1 inserts $U(1/4)$ and $U(3/4)$ (dashed boxes: newly inserted rungs);
Level~2 fills in all eighth-fractions. The closed set~$A$ ($f=0$) sits at the bottom and
$B$ ($f=1$) at the top. The vertical gap between adjacent rungs halves at each level;
after $k$~levels, the maximum gap is $2^{-k}$. Convergence occurs when the gap drops
below~$\varepsilon$.}
  \label{fig:ladder}
\end{subfigure}
\caption{From topology to dynamics. (a)~The static dyadic construction of the
Urysohn separator (\Cref{sec:foundation}): nested neighborhoods $\{P_r\}$ define
$f(x)=\inf\{r:x\in\mathrm{Int}\,P_r\}$ as a level-set infimum. (b)~The same
construction read dynamically as the \emph{Urysohn ladder} (\Cref{sec:dynamics}):
each dyadic level is one Evaluate--Detect--Refine round that inserts new rungs and
freezes existing ones, until the maximum gap falls below~$\varepsilon$. Reading
(a)\,$\to$\,(b) is the topology-to-dynamics transition: the static frontier system
of~(a) is exactly what the incremental process of~(b) lays down rung by rung.}
\Description{Two side-by-side panels. Left: nested dyadic neighborhoods separating
closed sets A and B, with frontiers shown as level boundaries. Right: a three-level
ladder inserting progressively finer rungs between A (value 0) and B (value 1).}
\label{fig:topology-to-dynamics}
\end{figure}

\paragraph{Connection to the Urysohn Machine}

The Fundamental Theorem provides the topological foundation for the~\UM{}'s construction
in three ways.
First, each Urysohn Triple $(\Sigma_c, \Pi_c, f_c)$ is one application of the theorem:
the support $\Sigma_c$ is the geometric realization of a subcomplex, the partition $\Pi_c$
defines the disjoint closed sets $A$ and $B$, and the classifier $f_c$ is the separator
constructed by the dyadic refinement. The theorem guarantees that $f_c$ exists and is
continuous, and the chain-level frontier calculus provides the algebraic structure that
the decision-boundary width measures: the total boundary measure
$\Width_{\partial}(\tau)$ of a triple $\tau$ is the
$(d{-}1)$-dimensional measure of the realized frontier support.  For a library or
multi-frontier realization, the corresponding Urysohn width $\Width_U$ is the sum
of these frontier measures.
Second, the dyadic refinement structure (Fig. \ref{fig:dyadic}) maps onto the~\UM{}'s stack discipline. Each
level of the dyadic family $\{P_r\}$ is a nested refinement of the previous level, just
as each triple pushed onto the stack refines the classification of the triple below it.
The inductive construction (T2: insert a midpoint between two existing levels) is the
algebraic analog of the~\UM{}'s context allocation: a new triple is inserted between
existing triples in the library.
Third, the algebraic coherence identity $\partial^2 = 0$ ensures that the frontier
calculus is self-consistent across composition. When two triples are composed sequentially, their frontiers interact; the coherence identity guarantees
that the composed frontier system remains a cycle system.Although Urysohn Width is specific to the present model, its subadditivity under
sequential composition is the chain-level analogue of standard facts from
algebraic topology and geometric measure theory \cite{hatcher2002algebraic,
munkres1984elements}: the boundary operator is linear,
\(\partial(c+c')=\partial c+\partial c'\), satisfies \(\partial^2=0\), and the
mass of a sum of chains/currents is subadditive~\cite{federer1969geometric,federer1960normal,ambrosio2000currents}.

\section{Decision Boundary Width and Amortized Separation Theorem}
\label{sec:width}

The~\UM{}'s principal theoretical contribution is the formalization of intrinsic
classification complexity in metric space.  To avoid ambiguity, we distinguish two
related but different quantities.  The first is the \emph{decision-boundary width}:
the geometric size of the boundary induced by a single classifier \cite{cover1965geometry}.  The second is the
\emph{Urysohn width}: the aggregate separation complexity of a Urysohn realization,
computed over the frontier system or over the library of Urysohn triples.  These
quantities coincide only in the special case where the realization consists of a
single boundary component.

\begin{definition}[Decision boundary]\label{def:boundary}
  Let \(\tau=(\Sigma,\Pi,f)\) be a Urysohn triple, where \(\Sigma\) is embedded in
  a \(d\)-dimensional metric space.  The \emph{decision boundary} of \(\tau\) is
    $\partial\tau
    :=
    \bigl\{x\in \Sigma:
    \text{every neighborhood of \(x\) contains points assigned to distinct labels}
    \bigr\}$.
  Equivalently, when \(f\) is regarded as a label-valued map, \(\partial\tau\)
  is the discontinuity set of \(f\).  If \(\partial\tau\) is sufficiently regular,
  it carries a natural \((d-1)\)-dimensional Hausdorff measure.
\end{definition}

\begin{definition}[Decision-boundary width]\label{def:boundary-width}
  The \emph{decision-boundary width} of a triple \(\tau\), denoted
  \(\Width_{\partial}(\tau)\), is the \((d-1)\)-dimensional measure of its
  decision boundary:
    $\Width_{\partial}(\tau)
    :=
    \mathcal H^{d-1}(\partial\tau)$.
  In dimension \(d=2\), this is boundary length; in dimension \(d=3\), it is
  boundary area; in general, it is codimension-one boundary measure.
\end{definition}

\begin{definition}[Urysohn width]\label{def:urysohn-width}
  Let
    $T=\{\tau_1,\ldots,\tau_k\}$
  be a finite Urysohn realization or a library of Urysohn triples.  The
  \emph{Urysohn width} of \(T\), denoted \(\Width_U(T)\), is the aggregate
  boundary complexity
    $\Width_U(T)
    :=
    \sum_{i=1}^{k}\Width_{\partial}(\tau_i)$.
  Equivalently, when a realization is represented by a finite chain-level
  frontier system \(\{\Gamma_\ell\}_{\ell\in \mathcal L}\), one may write
    $\Width_U(\mathcal R)
    :=
    \sum_{\ell\in\mathcal L}
    \mathcal H^{d-1}(|\Gamma_\ell|)$,
  where \(|\Gamma_\ell|\) denotes the geometric support of the frontier cycle.
  Thus decision-boundary width is a local geometric quantity attached to one
  separator, whereas Urysohn width is an amortized, realization-level quantity
  measuring the total frontier mass that the~\UM{} must store, compose, or reuse.
\end{definition}

Decision-boundary width captures the geometric difficulty of a single
classification boundary, while Urysohn width captures the total amount of
separation structure represented by a Urysohn library or frontier calculus.  For
example, a circular binary classifier in the plane has
  $\Width_{\partial}(\tau)=2\pi r $.
An Archimedean spiral classifier with \(n\) turns has boundary length growing as
\(\Theta(n^2)\), assuming the spiral radius grows linearly with the number of
turns (Fig.~\ref{fig:width-comparison}).  If these classifiers are each represented by a single Urysohn triple,
then \(\Width_U=\Width_{\partial}\).  If, however, the separator is decomposed
into multiple local triples, then \(\Width_U\) is the sum of the widths of those
local frontier components.  The distinction is important: \(\Width_{\partial}\)
measures the size of the target boundary, whereas \(\Width_U\) measures the
total boundary mass used by a particular Urysohn realization.
This distinction also clarifies what must be estimated from data. The library itself records the realized quantity \(\Width_U\), because its frontier components are explicitly represented. In contrast, the target quantity \(\Width_{\partial}\) is usually latent: from samples, the~\UM{} observes only the metric relationships among points and their class labels, not the continuous decision boundary itself. To make Width operational, we need an estimator that converts labeled metric data into a boundary-measure proxy. 

\paragraph{The contrastive separation operator and spectral estimation of Width}

Definitions~\ref{def:boundary}-\ref{def:urysohn-width} make decision-boundary width and Urysohn width well-defined geometric quantities, but they do not yet say how boundary width can be estimated from the data available to a~\UM{}, namely the metric and the class labels. A complexity measure that cannot be estimated is of limited operational value. We now show that \(\Width_{\partial}\) is recoverable, up to geometric normalization, from a graph-cut functional associated with a single class-aware operator, while the operator's spectrum provides complementary certificates of connectivity and conductance. 

\begin{definition}[Contrastive separation operator]\label{def:cs-operator}
  Let $(\Sigma,d)$ be the support of a triple $\tau=(\Sigma,\Pi,f)$ equipped with
  a sampling measure~$\mu$, and let $h>0$ be a scale parameter.  The
  \emph{contrastive separation (CS) operator} $\mathcal{L}_h$ acts on
  $L^2(\Sigma,\mu)$ by
  \[
    (\mathcal{L}_h u)(x)
    = \frac{1}{h^2}\int_{\Sigma}
      k_h(x,y)\,\bigl(u(x)-u(y)\bigr)\,d\mu(y),
    \qquad
    k_h(x,y)=\exp\!\Bigl(-\tfrac{d(x,y)^2}{h^2}\Bigr)\,
             \mathbf{1}\!\left[f(x)=f(y)\right].
  \]
  The kernel $k_h$ connects two points only when they carry the same label, so
  $\mathcal{L}_h$ is the Laplacian of the \emph{within-class} affinity graph.  Its
  spectrum is real and nonnegative, with eigenvalues
  $0=\lambda_0\le\lambda_1\le\lambda_2\le\cdots$, and the multiplicity of the
  eigenvalue~$0$ equals the number of class-connected components of~$\Sigma$.
\end{definition}

\paragraph{Variational (graph-cut) estimation of boundary width}

For sampled data, the geometric quantity \(\Width_{\partial}(\tau)\) is not
observed directly.  It can be estimated from a labeled neighborhood graph.  Given
samples \(x_1,\ldots,x_n\in\Sigma\) with labels \(y_i=f(x_i)\), construct a
weighted graph with kernel bandwidth \(h>0\):
  $W_{ij}
  =
  \eta\!\left(\frac{\|x_i-x_j\|}{h}\right)$,
where \(\eta\) is a compactly supported nonnegative kernel, and let
\(L=D-W\) be the graph Laplacian \cite{merris1994laplacian}.  For binary labels \(y_i\in\{0,1\}\), the
graph Dirichlet energy
  $y^\top L y
  =
  \frac12\sum_{i,j}W_{ij}(y_i-y_j)^2$
counts, with weights, the graph edges crossing the empirical decision boundary.
A normalized nonlocal-perimeter estimator is therefore
  $\widehat{\Width}_{\partial,h}(\tau)
  :=
  \frac{1}{c_{\eta,d}\,n^2h^{d+1}}
  \sum_{i,j}W_{ij}\mathbf 1[y_i\neq y_j]$,
where \(c_{\eta,d}\) is a kernel- and dimension-dependent normalization
constant.  Under standard sampling and regularity assumptions, this estimator
converges to a density-weighted version of
\(\mathcal H^{d-1}(\partial\tau)\).  If the sampling density is nonuniform, one
may use density-corrected weights, for example replacing \(W_{ij}\) by
\(W_{ij}/(\hat p_i\hat p_j)\), to estimate the unweighted Hausdorff measure more
faithfully.
Spectral quantities derived from \(L\), such as the second eigenvalue or the
Cheeger ratio \cite{chung1997spectral}, should be interpreted carefully.  They estimate
bottleneck or cut difficulty, not boundary width itself.  The direct estimator
of boundary width is the normalized cross-label graph cut above; Laplacian
eigenvalues provide complementary information about how strongly the boundary
separates the graph into class-consistent regions.
The cross-label cut above recovers the \emph{magnitude} of boundary measure.  The
contrastive operator $\mathcal{L}_h$ of \Cref{def:cs-operator} additionally
exposes the \emph{topology} of the class decomposition through its kernel and
low-lying spectrum.  These are distinct functionals of the same operator, and
only the cut estimates $\Width_{\partial}$; we now make both statements precise
and, in doing so, correct a tempting but false identification of Width with the
spectral gap.

\begin{theorem}[Consistency of the boundary-width estimator]\label{thm:cut-consistency}
  Fix the normalization constant
    $c_{\eta,d}=\sigma_\eta
    :=
    \int_{\R^{d}}\eta(|z|)\,\bigl|\langle z,e\rangle\bigr|\,dz$,
  which is independent of the unit vector \(e\) when \(\eta\) is radial.  Assume
  \textup{(A1)} the sampling density satisfies
  \(0<p_{\min}\le p\le p_{\max}\) on a bounded \(\Sigma\);
  \textup{(A2)} \(\partial\tau\) is a \(C^{1,1}\) hypersurface with finite
  measure \(\Width_{\partial}(\tau)=\Hdone(\partial\tau)\);
  \textup{(A3)} \(\eta\) is bounded, radial, nonnegative, compactly supported,
  and measurable; and
  \textup{(A4)} \(h=h_n\to0\) with
    $\frac{n h_n^{2d+2}}{\log n}\longrightarrow\infty$.
  Then, almost surely,
    $\widehat{\Width}_{\partial,h_n}
    \;\xrightarrow[n\to\infty]{}\;
    \int_{\partial\tau} p(x)^{2}\,d\Hdone(x)$.
  With density-corrected weights \(W_{ij}\mapsto W_{ij}/(\hat p_i\hat p_j)\), for a
  uniformly consistent density estimate \(\hat p\), the limit becomes the unweighted
  measure \(\Hdone(\partial\tau)=\Width_{\partial}(\tau)\).
\end{theorem}

\paragraph{What the spectrum certifies}

The eigenvalues of the within-class operator $\mathcal{L}_h$ report \emph{class
connectivity}, not boundary measure.  We record the three correct readings into the following proposition.

\begin{proposition}[Spectral certificates of the CS operator]\label{prop:cs-spectrum}
  Let \(G_h=(V,W)\) be the empirical within-class graph induced by
  \(\mathcal{L}_h\), let \(L_{\rm un}=D-W\) be its unnormalized Laplacian, and let
    $L_{\rm norm}=I-D^{-1/2}WD^{-1/2}$
  be the normalized Laplacian on vertices of positive degree.  Then:
  \begin{enumerate}[leftmargin=2em,nosep]
    \item \textup{(Topology / Hodge degree \(0\).)}
    \(\dim\ker L_{\rm un}\) equals the number of connected components of \(G_h\),
    equivalently the empirical degree-zero Betti number \(b_0(G_h)\).  This
    counts class-connected components but does not estimate their
    \((d{-}1)\)-dimensional boundary measure.

    \item \textup{(Conductance / Cheeger certificate.)}
    For any connected component \(H\) of \(G_h\), let \(\lambda_2(H)\) be the
    first nonzero eigenvalue of the normalized Laplacian of \(H\), and let
    \(\Phi(H)\) be its graph conductance.  Then
      $\frac{\lambda_2(H)}{2}\le \Phi(H)\le \sqrt{2\lambda_2(H)}$.
    Thus \(\lambda_2(H)\) certifies the internal bottleneck of a class component,
    not the measure of the decision boundary.

    \item \textup{(Stability of the count.)}
    If the first nonzero spectral gap
    \(\gamma=\lambda_{b_0+1}(L_{\rm norm})-\lambda_{b_0}(L_{\rm norm})\) is
    positive and the graph Laplacian is perturbed by an operator-norm amount
    \(\delta<\gamma/2\), then the spectral projector onto the empirical
    zero-eigenspace changes by \(O(\delta/\gamma)\).  Hence the reliability of
    the component count and of the induced near-partition is governed by this
    eigengap, via the Davis-Kahan theorem \cite{davis1970rotation}.
  \end{enumerate}
\end{proposition}

\begin{remark}[The spectral gap does not estimate boundary width]\label{rem:no-spectral-width}
  The conductance probed by $\lambda_1$ (Proposition \ref{prop:cs-spectrum}(2)) is an
  isoperimetric \emph{ratio} of a class region's interior and is independent of
  $\Width_{\partial}$.  In $\Sigma=[0,1]^2$, take (a) a thin strip
  $[0,1]\times[0,\delta]$ as one class: its boundary length is $\Theta(1)$ but,
  being path-like, its within-class gap is $\lambda_1=\Theta(\delta^{2})\to0$;
  versus (b) a fixed-radius disk whose boundary is folded into $m$ ripples of
  fixed amplitude: its boundary length is $\Theta(m)\to\infty$ while its interior
  stays well connected, $\lambda_1=\Theta(1)$.  Small and large $\Width_{\partial}$
  are compatible with small and large $\lambda_1$ in all four combinations, so
  $\lambda_1$ is not a function of $\Width_{\partial}$.  Boundary width is
  estimated by \Cref{thm:cut-consistency}; the spectrum is reserved for the
  topological and conductance certificates of Proposition \ref{prop:cs-spectrum}.  
\end{remark}

\paragraph{The Amortized Separation Theorem}

The preceding realization theorem shows that a Urysohn separator can be
represented by a coherent dyadic frontier calculus.  The next result asks a
complementary approximation-theoretic question: how many simple separating
components are required to reproduce a separator whose boundary has nontrivial
decision boundary width?  Intuitively, a decision boundary with large
\(\Width_{\partial}(\tau)\) cannot be captured by a small number of localized basis
functions, because each simple basis triple can resolve only a bounded amount of
separation complexity at accuracy \(\eps\).  Therefore, decision boundary width acts as an
intrinsic obstruction to compression: the larger the width of the boundary, the
larger the number of basis functions required to approximate the corresponding
separator \cite{devore1998nonlinear,devore1993constructive}.  The theorem formalizes this obstruction as an amortized lower bound,
showing that separation complexity cannot be hidden by reuse of simple local
components unless the number of such components grows at least linearly with
\(\Width_{\partial}(\tau)\) and inversely with the target accuracy.

\begin{theorem}[Amortized Separation Theorem]\label{thm:amortized}
  Let \(\tau=(\Sigma,\Pi,f)\) be a Urysohn triple whose decision boundary has
  finite width \(\Width_{\partial}(\tau)\).  Let
  \(\{B_1,\ldots,B_k\}\) be \(k\) basis triples with simple supports.  Suppose
  that an \(\eps\)-accurate approximation of \(f\) requires the boundary
  \(\partial\tau\) to be covered by the boundary footprints \(U_i\) of these
  supports, and suppose each simple support resolves at most \(C\eps\) units of
  \((d-1)\)-dimensional boundary measure:
    $\Hdone(\partial\tau\cap U_i)\le C\eps
    \quad\text{for every }i$.
  Then
    $k \;\geq\; \frac{\Width_{\partial}(\tau)}{C\eps}$,
  where \(C\) depends only on the ambient geometry and on the Lipschitz and
  regularity constants of the basis triples.
\end{theorem}

\begin{remark}[Estimability and the amortized bound]\label{rem:cs-amortized}
  \Cref{thm:cut-consistency} supplies the quantity that drives the lower bound of
  the next subsection: because $\Width_{\partial}(\tau)$ is consistently estimable
  from labeled metric data, the amortized triple count
  $k\ge\Width_{\partial}(\tau)/(C\eps)$ of \Cref{thm:amortized} is itself
  estimable, closing the loop between the complexity measure and the resource
  bound without any spurious spectral identity.  Finite-sample rates for
  $\widehat{\Width}_{\partial,h}$ are left to future work.
\end{remark}

The theorem has a natural interpretation: complex boundaries require proportionally more
computational resources, and no amount of clever programming can circumvent this lower
bound, which is the formal expression of intrinsic complexity \cite{vapnik1998statistical}. The lower bound, however, only describes the cost of representing a separator
from simple local components.  It does not explain how inference should move
once such a separator has been realized.  The next step carries an important intuition about going metric \cite{gromov1999metric}:
after separation has identified which points belong together and which points
must remain apart, inference should contract distances within the same class
while preserving, or even enlarging, distances across different classes.  In this
sense, Urysohn realization supplies the separating structure, while \emph{metric
contraction} supplies the computational geometry that makes inference efficient.

\begin{figure}[h]
\centering
\begin{subfigure}[b]{0.48\textwidth}
  \centering
  \resizebox{\linewidth}{!}{%
\begin{tikzpicture}[scale=0.8]
\begin{scope}[shift={(-4,0)}]
  \node[font=\small\bfseries] at (0,2.5) {Circle boundary};
  \fill[orange!15] (-2.2,-2.2) rectangle (2.2,2.2);
  \fill[blue!15] (0,0) circle (1.5);
  \draw[red, very thick] (0,0) circle (1.5);
  \node[font=\scriptsize] at (0,0) {class $A$};
  \node[font=\scriptsize] at (1.7,1.7) {class $B$};
  \node[font=\scriptsize, red!70!black] at (0,-2.7)
    {$\Width_{\partial}(\tau)=2\pi r$};
\end{scope}

\begin{scope}[shift={(4,0)}]
  \node[font=\small\bfseries] at (0,2.5) {Spiral boundary};
  \fill[orange!15] (-2.2,-2.2) rectangle (2.2,2.2);
  \draw[red, very thick, domain=0:1080, samples=300, smooth]
    plot ({1.8*(1-\x/1200)*cos(\x)}, {1.8*(1-\x/1200)*sin(\x)});
  \node[font=\scriptsize] at (-0.3,0.4) {$A$};
  \node[font=\scriptsize] at (1.5,-1.5) {$B$};
  \node[font=\scriptsize, red!70!black] at (0,-2.7)
    {$\Width_{\partial}(\tau)=\Theta(n^2)$};
\end{scope}

\draw[-{Stealth[length=5pt]}, thick, gray!50] (-1.0,0) -- (1.0,0)
  node[midway, above, font=\scriptsize\itshape, text=gray] {larger boundary width};
\end{tikzpicture}}
  \caption{Decision-boundary width versus realized Urysohn width.  
  If each classifier is represented by a single Urysohn triple, then the realized Urysohn width satisfies
\(\Width_U=\Width_{\partial}\).  More generally, if the separator is decomposed
into multiple local triples, then \(\Width_U\) is the sum of the widths of all
realized frontier components.  
}
  \label{fig:width-comparison}
\end{subfigure}\hfill
\begin{subfigure}[b]{0.48\textwidth}
  \centering
  \resizebox{\linewidth}{!}{%
\begin{tikzpicture}[scale=0.85]
\begin{scope}[shift={(-4.5,0)}]
  \node[font=\small\bfseries] at (0,2.5) {Before contraction};
  \fill[blue!15] (-2,-2) rectangle (2,2);
  \fill[red!15] plot[smooth, tension=0.7] coordinates
    {(-2,0.3) (-1,0.8) (0,-0.3) (1,0.6) (2,0.1)} -- (2,2) -- (-2,2) -- cycle;
  \draw[thick, red!60!black, smooth, tension=0.7]
    plot coordinates {(-2,0.3) (-1,0.8) (0,-0.3) (1,0.6) (2,0.1)};
  \node[font=\scriptsize] at (0,1.4) {class $A$};
  \node[font=\scriptsize] at (0,-1.5) {class $B$};
  \fill (0.5,-0.8) circle (2pt) node[right, font=\tiny] {$x$};
  \fill (-0.5,-0.6) circle (2pt) node[left, font=\tiny] {$x'$};
  \draw[{Stealth[length=3pt]}-{Stealth[length=3pt]}, gray, thick]
    (0.5,-0.8) -- node[below, font=\tiny] {$d_X(x,x')$} (-0.5,-0.6);
  \node[font=\scriptsize, red!60!black] at (0,-2.3) {$\Width = $ long boundary};
\end{scope}
\begin{scope}[shift={(4.5,0)}]
  \node[font=\small\bfseries] at (0,2.5) {After contraction $G_c$};
  \fill[blue!15] (-2,-2) rectangle (2,2);
  \fill[red!15] (-2,0.8) rectangle (2,2);
  \draw[thick, red!60!black] (-2,0.8) -- (2,0.8);
  \node[font=\scriptsize] at (0,1.4) {class $A$};
  \node[font=\scriptsize] at (0,-1.4) {class $B$};
  \fill (0.2,-0.3) circle (2pt) node[right, font=\tiny] {$G_c(x)$};
  \fill (-0.1,-0.2) circle (2pt) node[left, font=\tiny] {$G_c(x')$};
  \draw[{Stealth[length=3pt]}-{Stealth[length=3pt]}, gray, thick]
    (0.2,-0.3) -- node[below, font=\tiny] {$\kappa\!\cdot\! d_X$} (-0.1,-0.2);
  \node[font=\scriptsize, red!60!black] at (0,-2.3) {$\lambda \!\cdot\! \Width < \Width$};
\end{scope}
\draw[-{Stealth[length=6pt]}, very thick, gray!50] (-1.5,0) -- (1.5,0)
  node[midway, above, font=\scriptsize\itshape, text=gray] {$G_c$: contract};
\end{tikzpicture}}
  \caption{Metric contraction reshapes the input space so that within-class distances shrink
($d_Z \leq \kappa \cdot d_X$, $\kappa < 1$) and the decision boundary straightens. The
Geodesic Inference Bound guarantees that the contracted boundary length
$\lambda \cdot \Width$ is strictly shorter than the original~$\Width$, reducing inference
to a short geodesic traversal rather than ambient-space search.}
  \label{fig:contraction}
\end{subfigure}
\caption{Width and its reduction by metric contraction. (a)~\emph{The cost.}
Decision-boundary width measures the boundary a separator must resolve: a circular
classifier has $\Width_{\partial}=2\pi r$, whereas a spiral with $n$ turns grows as
$\Theta(n^2)$. (b)~\emph{The mechanism.} Metric contraction $G_c$ shrinks
within-class distances ($d_Z\le\kappa\,d_X$, $\kappa<1$) so the same partition is
resolved along a boundary of length $\lambda\,\Width<\Width$. Panel~(a) fixes the
geometric cost a separator must pay; panel~(b) is the mechanism that lowers the
\emph{effective} cost paid at inference time.}
\Description{Two side-by-side panels. Left: a circular decision boundary of length
2 pi r beside a spiral boundary of much greater length. Right: a before-and-after
diagram in which metric contraction reduces within-class distances and straightens
the decision boundary.}
\label{fig:width-and-contraction}
\end{figure}

\paragraph{Metric contraction and geodesic inference}

A key consequence of the~\UM{} formalism is that once a separator has been
constructed, inference need not be performed by repeatedly solving the original
separation problem.  Instead, the separator can be reused as a geometric
shortcut (the metric analogue of amortized inference
\cite{gershman_amortized_2014}): an expensive separation structure is built once,
stored in the Metric Library, and then reused to make subsequent inference paths
shorter.
The Urysohn separator identifies which regions of the space should be treated as
distinct, but efficient inference also requires a geometry in which movement
within an already resolved region is cheaper than movement across an unresolved
boundary.  Separation and contraction therefore play complementary roles.
Separation creates the class-level partition, while contraction reshapes the
metric so that points belonging to the same class become closer and points
belonging to different classes remain separated.  In this way, the~\UM{} does
not store a classifier but a modified geometry in which future
queries can be routed through class-consistent regions.

The Urysohn viewpoint turns classification into a geodesic problem.  Once a separator
has been realized, an inference procedure should not search the ambient space
uniformly.  It should instead follow short paths inside class-consistent regions
and avoid paths that cross committed decision boundaries.  The separator
induces an anisotropic computational geometry: directions compatible with the
current class are contracted, whereas directions leading across class boundaries
are preserved or expanded.  The resulting geodesics implement iterative amortization \cite{marino_iterative_2018}:
future queries reuse the contracted geometry rather than reconstructing the
separator from scratch.  The following definition formalizes this idea by
introducing a metric that is contractive within each class and noncontractive
across different classes.

\begin{definition}[Class-aware contraction]\label{def:contraction}
  Let $(S,d)$ be a metric space with a classification $f\colon S \to \Pi$.  A
  \emph{class-aware contraction} is a metric
  $d'\colon S \times S \to \R_{\geq 0}$ such that:
  \begin{enumerate}
    \item $d'(x,y) \leq \lambda d(x,y)$ for all $x,y$ with
      $f(x)=f(y)$, where $\lambda<1$; and
    \item $d'(x,y) \geq d(x,y)$ for all $x,y$ with $f(x)\neq f(y)$.
  \end{enumerate}
\end{definition}

Under class-aware contraction, the topology of class regions is preserved while
the metric cost of moving inside each region is reduced.  Consequently, the
geodesic structure changes: paths that remain within a single class become
shorter, whereas paths that cross class boundaries become relatively longer.
This is precisely the mechanism by which the Metric Library amortizes inference.
The library does not eliminate the cost of constructing the separator; rather,
it converts that one-time cost into a reusable geometry for subsequent queries.

\begin{theorem}[Geodesic Amortization Bound]\label{thm:geodesic}
  Let \(\Lib\) be a Metric Library with stack depth \(d\).  Suppose an inference
  path decomposes into level-wise portions of boundary mass
  \(W_1,\ldots,W_d\), with
    $\sum_{i=1}^{d} W_i \leq \Width_U(\Lib)$,
  and suppose that level \(i\) contracts within-class path length by a factor
  \(\lambda_i<1\).  Then the contracted inference path length is at most
    $\sum_{i=1}^{d}\lambda_i W_i$.
  In particular, if the boundary mass is equidistributed across levels, so that
  \(W_i\leq \Width_U(\Lib)/d\), then the contracted path length is at most
    $\sum_{i=1}^{d}\lambda_i\,\frac{\Width_U(\Lib)}{d}$.
  For uniform contraction \(\lambda_i=\lambda<1\), this bound becomes
    $\lambda\,\Width_U(\Lib)<\Width_U(\Lib)$.
\end{theorem}

The~\UM{} stack gives a discrete realization of the amorization principle.  Each level of
the stack represents a scale at which part of the separation structure has
already been resolved and can therefore be reused.  Lower levels provide coarse
class-consistent contractions, while higher levels refine the geometry near
finer decision boundaries.  Inference proceeds by following the contracted
geodesic structure induced by the stack, rather than by searching through the
ambient space or reconstructing the full boundary of width~\(\Width_U(\Lib)\).
In summary, metric contraction supplies the geometric mechanism for amortized
inference \cite{gershman_amortized_2014}: the cost of building the frontier calculus is paid during
realization, while later queries benefit from shorter class-aware paths
(Fig.~\ref{fig:contraction}).


\section{From Topology to Dynamics: The Urysohn Ladder as Computation}\label{sec:dynamics}
 
Sections~\ref{sec:um}-\ref{sec:width} establish the~\UM{}'s static structure: what the
Metric Library stores and what it costs. This section derives the~\UM{}'s \emph{dynamics}
from that topology, showing that the incremental construction of a separator, the optimal
library construction strategy, and the mechanism of generalization all follow from the
Urysohn ladder construction (Fig. \ref{fig:ladder}).

\paragraph{Incremental construction along the Urysohn ladder}

The proof of the Fundamental Theorem (\Cref{thm:realization}) constructs the separator
$f$ by iteratively inserting nested open sets $\{U(r)\}_{r \in \mathbb{D}}$. This dyadic
refinement is not only a proof technique but also the procedure by which a~\UM{} builds a
separator (Fig. \ref{fig:log_scaling}). Each insertion step is one round of the construction:
\begin{itemize}[nosep]
  \item \textbf{Evaluate}: Apply the current partial separator $f_k$ (defined
    from the $k$~existing rungs) to classify an input $x$. The value
    $f_k(x) = \inf\{r : x \in U(r)\}$ is well-defined for the existing rungs.
  \item \textbf{Detect}: Identify a pair of adjacent rungs $U(r) \Subset U(s)$
    between which the classification is too coarse - the gap $|s - r|$ exceeds the
    required resolution $\varepsilon$. A new input that falls between existing rungs
    and cannot be classified at the required precision signals that the gap is too wide.
  \item \textbf{Refine}: Insert a new rung $U(t)$ with $r < t < s$ and
    $U(r) \Subset U(t) \Subset U(s)$ (using Axiom T2 or T3 from
    \Cref{def:dyadic-normality}). The existing rungs are frozen: the nesting condition
    $\overline{U(r)} \subset U(t)$ is permanent. The library grows by one triple; the
    separator $f_{k+1}$ is strictly finer than $f_k$.
\end{itemize}

At convergence, the dyadic family is dense in $[0,1]$, the separator is continuous, and
no further insertions are needed. This is the~\UM{}'s convergence condition:
$|\M| = \mathrm{uw}(\mathcal{P})$ means the ladder has enough rungs.

\begin{proposition}[Convergence of ladder refinement]\label{prop:edt-convergence}
  Let $\varepsilon > 0$ and let $\{U(r)\}$ be a dyadic family constructed by incremental
  refinement. Suppose that a full refinement round subdivides every active label gap by
  a factor at least $\rho>1$. After at most
  $\lceil \log_\rho(1/\varepsilon) \rceil$ full refinement rounds, every unresolved
  label gap has width at most $\varepsilon$, so no further refinement is triggered at
  resolution~$\varepsilon$. If a refinement event refines only one gap at a time, then the
  same guarantee requires enough events to refine all active gaps; the logarithmic bound
  applies to rounds, not individual single-gap insertions.
\end{proposition}

\paragraph{Optimal radix economy}
 
The standard Urysohn proof uses $\rho = 2$ (binary subdivision) because the proof
requires only existence, not optimality. The~\UM{} asks: which base~$\rho$ minimizes the
total cost of building the ladder?
At each level, the~\UM{} must maintain a number of simultaneous alternatives
proportional to the radix~$\rho$ (the ``width cost''). The number of levels needed to
achieve resolution~$\varepsilon$ is
$\log_\rho(1/\varepsilon)=\ln(1/\varepsilon)/\ln\rho$ (the ``depth cost''). Then the
standard radix-economy cost is
$\mathcal{C}(\rho)=\rho\,\frac{\ln(1/\varepsilon)}{\ln\rho}$.
Treating $\rho$ as a continuous variable, this cost is minimized at $\rho=e$.  If the
radix must be an integer, the optimal choice is the nearest integer radix, namely~$3$
\cite{hayes2001third}.
 
\begin{corollary}[Optimal Urysohn ladder]\label{cor:optimal-radix}
  Under the standard radix-economy cost
  $\mathcal{C}(\rho)=\rho\ln(1/\varepsilon)/\ln\rho$, the continuous optimum is
  $\rho=e$. The optimal depth is $\ln(1/\varepsilon)$ and the optimal continuous-radix
  cost is $e\ln(1/\varepsilon)$. For integer radix, the optimum is attained at $\rho=3$.
\end{corollary}

The above result connects the decision boundary Width (the total boundary length, a static quantity) to
the optimal construction strategy (the number of refinement rounds, a dynamic quantity).
The Width determines \emph{how much} boundary must be resolved; the radix economy
determines \emph{how efficiently} the ladder resolves it.

\begin{remark}[Stack depth is the space resource: a Savitch-style economy]
\label{rem:stack-savitch}
  The dyadic refinement of \Cref{thm:realization} is recursive midpoint
  divide-and-conquer: separating at resolution $\rho^{-k}$ inserts a level between
  two existing levels and recurses on each side. This is the recursion structure
  of Savitch's space-reuse argument \cite{savitch1970relationships}, and it has the same
  consequence here. A~\UM{} that builds the separator depth-first holds only the
  chain of nested supports from the root to the current level on its stack, so its
  peak space is the stack \emph{depth}, $O(\log_\rho(1/\varepsilon))$, exactly the
  depth cost of \Cref{cor:optimal-radix}. A breadth-first realization (the natural
  queue-based alternative) would instead hold an entire level's frontier system
  simultaneously and so would consume space proportional to the realized
  \emph{width}, $\Omega(\Width)$. The last-in-first-out discipline is therefore
  not only the topologically natural one but also the space-economical one: it
  trades the width of the frontier for the depth of the ladder, reusing the support
  region as it descends. We state this as a structural correspondence rather than a
  complexity-class theorem; turning it into a formal characterization of a
  space-bounded~\UM{} class, with
  stack depth as the space measure, would make the Savitch correspondence precise,
  and we leave it to future work.
\end{remark}

\begin{figure}[h]
\centering
\resizebox{\linewidth}{!}{%
\begin{tikzpicture}[
    scale=0.9, transform shape,
    manifold/.style={
        draw=gray!50, 
        fill=gray!10, 
        rounded corners, 
        minimum height=1.2cm, 
        inner sep=5pt,
        align=center
    },
    traj/.style={
        thick, 
        decorate, 
        decoration={snake, amplitude=1pt, segment length=5pt},
        color=blue!70
    },
    condensed_traj/.style={
        very thick,
        color=blue!90!black,
        smooth
    },
    raw_point/.style={circle, fill=blue!50, inner sep=1pt},
    condensed_point/.style={diamond, fill=red!70, inner sep=2pt},
    map_arrow/.style={->, -{Stealth[scale=1.2]}, thick, color=red!80!black},
    contraction_cone/.style={
        fill=red!30, 
        opacity=0.4, 
        draw=none
    },
    dim_line/.style={
        |<->|, 
        thick, 
        >=Stealth,
        color=black!70
    },
    odd_phase/.style={
        fill=orange!20,
        draw=orange!60!black,
        rounded corners=3pt,
        thick,
        minimum width=2.8cm,
        minimum height=0.9cm,
        align=center,
        font=\footnotesize
    },
    even_phase/.style={
        fill=blue!15,
        draw=blue!60!black,
        rounded corners=3pt,
        thick,
        minimum width=2.8cm,
        minimum height=0.9cm,
        align=center,
        font=\footnotesize
    },
    pump_arrow/.style={
        ->, -{Stealth[scale=1]},
        very thick
    }
]
 
 
\node[manifold, minimum width=12cm] (M0) at (0,0) {};
\node[below right, gray] at (M0.north west) {$\mathcal{M}_0$ (Base Manifold)};
\draw[thick, red!70, smooth] (-5.2, 0.15) to[out=20,in=160] (-3, -0.1) to[out=-20,in=200] (-1, 0.2) to[out=20,in=160] (1, -0.15) to[out=-20,in=200] (3, 0.1) to[out=20,in=160] (5.2, -0.1);
\draw[thick, blue!70, smooth] (-5.2, -0.15) to[out=-20,in=200] (-3, 0.1) to[out=20,in=160] (-1, -0.2) to[out=-20,in=200] (1, 0.15) to[out=20,in=160] (3, -0.1) to[out=-20,in=200] (5.2, 0.1);
\node at (0, -0.45) {\footnotesize Entangled class manifolds ($\delta \approx 0$)};
\draw[dim_line] (-5.5, -0.9) -- (5.5, -0.9) node[midway, below] {Linear Diameter $\propto L$};
 
\draw[<->, orange!80!black, thick] (5.8, -0.15) -- (5.8, 0.15);
\node[right, orange!80!black, font=\scriptsize] at (5.9, 0) {$\delta_0 \!\approx\! 0$};

\node[manifold, minimum width=7cm] (M1) at (0, 3.5) {};
\node[below right, gray] at (M1.north west) {$\mathcal{M}_1$ (Quotient Manifold)};
\coordinate (c1_1) at (-2.5, 3.7);
\coordinate (c1_2) at (0, 3.3);
\coordinate (c1_3) at (2.5, 3.7);
\fill[red!70] (-2.5, 3.7) circle (3pt);
\fill[red!70] (2.5, 3.7) circle (3pt);
\fill[blue!70] (0, 3.3) circle (3pt);
\draw[condensed_traj, red!50] (-2.5, 3.7) to[out=-10, in=190] (2.5, 3.7);
\draw[condensed_traj, blue!50] (-1.2, 3.3) -- (1.2, 3.3);
 
\draw[<->, green!60!black, thick] (3.1, 3.3) -- (3.1, 3.7);
\node[right, green!60!black, font=\scriptsize] at (3.2, 3.5) {$\delta_1 > 0$};

\begin{pgfonlayer}{background}
    \fill[contraction_cone] (-5.5, 0.3) -- (-1.5, 0.3) -- (c1_1) -- cycle;
    \fill[blue!15, opacity=0.4, draw=none] (-1.5, 0.3) -- (1.5, 0.3) -- (c1_2) -- cycle;
    \fill[contraction_cone] (1.5, 0.3) -- (5.5, 0.3) -- (c1_3) -- cycle;
\end{pgfonlayer}

\node at (0, 5.2) {\Huge $\vdots$};
\node[right, gray, font=\footnotesize, align=left] at (1, 5.2) {Iterated\\pump cycles};

\node[manifold, minimum width=3cm] (MD) at (0, 7) {};
\node[below right, gray] at (MD.north west) {$\mathcal{M}_D$ (Top Manifold)};
\fill[red!90] (-0.8, 6.7) circle (4pt);
\fill[blue!90] (0.8, 6.7) circle (4pt);
\node[red!80!black, font=\scriptsize, above] at (-0.8, 6.75) {$A$};
\node[blue!80!black, font=\scriptsize, above] at (0.8, 6.75) {$B$};
\draw[green!60!black, very thick, dashed] (0, 6.5) -- (0, 7.5);
\node[green!60!black, font=\scriptsize, below] at (0, 6.45) {$f$};
 
\draw[dim_line] (-1.5, 7.8) -- (1.5, 7.8) node[midway, above, align=center] {Bounded Diameter\\$\approx L / \rho^D = O(1)$};
 
\draw[<->, green!60!black, very thick] (1.4, 7) -- (1.8, 7);
\draw[<->, green!60!black, very thick] (-1.4, 7) -- (-1.8, 7);
\node[right, green!60!black, font=\scriptsize] at (2.0, 7) {$\delta_D \gg 0$};
 
\begin{pgfonlayer}{background}
     \fill[contraction_cone] (M1.north west) -- (M1.north east) -- (MD.south) -- cycle;
\end{pgfonlayer}
\node[map_arrow, rotate=90] at (-2, 5.5) {$q_{D-1}$};

\draw[dim_line] (-7.5, 0) -- (-7.5, 7) node[midway, fill=white, align=center, rotate=90] {Hierarchy Depth $D = O(\log L)$};

 
\node[font=\bfseries, align=center] at (9, 7.8) {Separator Pump\\[-2pt]\footnotesize (Parity Alternation)};
 
 
\node[odd_phase] (odd1) at (9, 1.0) {\textbf{Odd} ($\Hodd$)\\[-2pt]\scriptsize Expand / Explore};
\draw[pump_arrow, orange!70!black] (odd1.south) ++(0,-0.05) -- ++(0, -0.5) node[right, font=\scriptsize, orange!70!black, pos=0.5] {expose $\partial$};
 
\node[even_phase] (even1) at (9, 2.5) {\textbf{Even} ($\Heven$)\\[-2pt]\scriptsize Contract / Validate};
\draw[pump_arrow, blue!60!black] (even1.south) ++(0,-0.05) -- ++(0, -0.5) node[right, font=\scriptsize, blue!60!black, pos=0.5] {collapse $\beta_1 \!\to\! \beta_0$};
 
 
\node[draw=green!60!black, fill=green!5, rounded corners=2pt, font=\scriptsize, align=center, minimum width=2.4cm] (delta1) at (9, 3.6) {$\delta_0 \to \delta_1$\\margin increases};

\draw[gray!40, thick, dashed] (7.2, 4.3) -- (10.8, 4.3);
 
 
\node[odd_phase] (odd2) at (9, 4.9) {\textbf{Odd} ($\Hodd$)\\[-2pt]\scriptsize Expand / Explore};
\draw[pump_arrow, orange!70!black] (odd2.south) ++(0,-0.05) -- ++(0, -0.5);
 
\node[even_phase] (even2) at (9, 6.2) {\textbf{Even} ($\Heven$)\\[-2pt]\scriptsize Contract / Validate};
\draw[pump_arrow, blue!60!black] (even2.south) ++(0,-0.05) -- ++(0, -0.5) node[right, font=\scriptsize, blue!60!black, pos=0.5] {collapse};
 
 
\node[draw=green!60!black, fill=green!5, rounded corners=2pt, font=\scriptsize, align=center, minimum width=2.4cm] (delta2) at (9, 7.1) {$\delta_{D-1} \to \delta_D$\\Urysohn activates};

 
\draw[->, thick, gray!50, dashed] (odd1.west) -- (5.7, 0.5) node[midway, above, font=\scriptsize, gray, sloped] {identifies};
\draw[->, thick, gray!50, dashed] (even1.west) -- (5.7, 2.8) node[midway, above, font=\scriptsize, gray, sloped] {collapses};
 
\node[align=center, font=\footnotesize, color=red!70!black] at (-8.5, 3.5) {\textbf{Geometric}\\\textbf{Necessity:}\\Trading linear\\width for\\logarithmic\\depth.};
 
\node[draw=black!40, fill=white, rounded corners=3pt, font=\footnotesize, align=center, text width=5cm] at (9, -0.8) {\textbf{Key:} Each odd-even cycle is one\\pump stroke. Part~1 (contraction) is\\the inner operation of Part~2 (alternation).};
 
\end{tikzpicture}
}
\caption{\textbf{The Urysohn Ladder with Separtor Pump.} \textbf{Left:} A hierarchy of quotient maps $q_k$ progressively compresses the input stream from $\mathcal{M}_0$ to $\mathcal{M}_D$, trading linear diameter for logarithmic depth. At $\mathcal{M}_0$, class manifolds (red, blue) are entangled with margin $\delta_0 \approx 0$. At $\mathcal{M}_D$, classes occupy well-separated tokens with $\delta_D \gg 0$, and Urysohn's separating function $f$ (green dashed line) exists with finite Lipschitz constant. \textbf{Right:} Each level transition is driven by the \emph{separator pump}: a two-stroke parity alternation cycle. The \emph{odd phase} ($\Hodd$, orange) expands the representation to expose boundary failures; the \emph{even phase} ($\Heven$, blue) contracts validated regions into tokens via metric contraction ($\beta_1 \to \beta_0$). Each cycle monotonically increases the margin $\delta$, iteratively manufacturing the disjointness precondition that Urysohn's Lemma requires.}
\label{fig:log_scaling}
\end{figure}

\paragraph{Abstraction and hallucination as class-aware contraction}

The contraction $d_Z(G(x), G(x')) \leq \kappa \cdot d_X(x, x')$
(\Cref{def:contraction}) has a geometric interpretation that bridges the static
width bounds to the dynamic refinement process. In any Riemannian manifold \cite{lee_introduction_2011}, the geodesic
distance is bounded below by the chordal (Euclidean) distance:
$d_{\mathrm{Eucl}}(x, y) \leq d_{\mathrm{geo}}(x, y)$.
The class-aware contraction $G_c$ acts as a \emph{relaxation operator} \cite{chung1997spectral}: it deforms the manifold's
metric to bring geodesic distances closer to the Euclidean lower bound, flattening the
local curvature and simplifying the optimization landscape.
 
\begin{itemize}[nosep]
  \item \textbf{Valid contraction (abstraction):} Collapsing within-class geodesic
    distances (e.g., ``poodle'' $\to$ ``dog'') is topologically safe because the points
    lie in the same basin of the Urysohn separator. The contraction does not cross any
    frontier $\Gamma_r$.
\item \textbf{Valid extension (generalization):} Assigning a new or nearby point (e.g., an unseen breed $\to$ ``dog'') to an existing class basin is safe when its shortest class-aware path attaches to that basin without crossing a committed frontier~$\Gamma_r$. Generalization therefore extends the contracted region along class-consistent geodesics rather than collapsing distinct basins.
  \item \textbf{Invalid contraction (hallucination):} Collapsing across-class geodesic
    distances (e.g., ``dog'' $\to$ ``cat'') tunnels through a frontier~$\Gamma_r$,
    conflating two regions that the separator is designed to distinguish.
\end{itemize}
 
The~\UM{}'s topological constraints, the frozen Urysohn Triples, the nesting condition
$\overline{U(r)} \subset U(s)$, the algebraic coherence $\partial^2 = 0$, collectively
prevent invalid contraction. The refinement step commits frontiers
permanently, ensuring that no subsequent contraction can tunnel through a committed
boundary. This is the sense in which the~\UM{} controls metric collapse: it permits
collapse within basins (improving efficiency) while forbidding collapse across basins
(preserving correctness).
 
\begin{remark}[Generalization as Tietze extension]
 
The Tietze Extension Theorem (a corollary of Urysohn's Lemma) states \cite{tietze_uber_1915}: any continuous
function defined on a closed subset of a normal space extends continuously to the
entire space. In the~\UM{}, this provides the mechanism of \emph{generalization}.
A classifier $f_c$ trained on a closed training set $\mathcal{M}_{\mathrm{train}} \subset
X$ extends to a continuous function on all of~$X$. The extension is mathematically
guaranteed to exist and to preserve the classification structure on~$\mathcal{M}_{\mathrm{train}}$.
However, the extension is reliable only within the region where the Urysohn ladder has
been calibrated:
$\mathcal{V}_\delta = \{x \in X : d_X(x, \mathcal{M}_{\mathrm{train}}) < \delta\}$,
where $\delta$ is the ladder's resolution scale. Inside~$\mathcal{V}_\delta$, the Tietze
extension is a valid generalization. Outside~$\mathcal{V}_\delta$, the function still
returns values (the model always produces outputs), but these values are extrapolations
beyond the domain of valid calibration (often known as hallucination \cite{ji2023survey}, as we will elaborate next).
\end{remark}

\begin{remark}[Hallucination as Tietze extension beyond the valid domain]
  In the~\UM{} framework, \emph{hallucination} is a domain-calibration failure:
  the Tietze extension is applied to inputs for which no reliable Urysohn ladder
  has been constructed.  The model therefore predicts that hallucination risk
  should increase with
  \(d_X(x,\mathcal M_{\mathrm{train}})\), the distance from the input to the
  training manifold in the library metric.  This view is consistent with
  empirical and theoretical work on distance-aware uncertainty: Liu
  et al.~\cite{liu2020sngp} show that distance awareness is necessary for
  minimax-optimal uncertainty estimation and that standard DNN features can fail
  by collapsing out-of-distribution points toward in-distribution regions; Lee
  et al.~\cite{lee2018mahalanobis} use Mahalanobis distance to class-conditional
  feature distributions for out-of-distribution detection; and Sun
  et al.~\cite{sun2022knn} show that \(k\)-nearest-neighbor distance in feature
  space detects unreliable predictions.  The difference between generalization and hallucination is not whether the
model produces an output, but whether the output is supported by a
frontier-preserving path back to the calibrated training domain.  Generalization
extends within a basin; hallucination jumps across, or beyond, the basin
structure.
\end{remark}

\section{Computational Guarantees of the Urysohn Ladder}\label{sec:guarantees}

Section~\ref{sec:dynamics} constructed the Urysohn ladder as a recursive
quotient computation: a tower
  $\mathcal M_0 \xrightarrow{\,q_0\,} \mathcal M_1 \xrightarrow{\,q_1\,}
  \cdots \xrightarrow{\,q_{D-1}\,} \mathcal M_D$,
in which each map $q_k$ collapses a committed (frozen) region of $\mathcal M_k$ to
a single point (a \emph{token}) of $\mathcal M_{k+1}$.  This is the topological
content of the freezing discipline of \Cref{def:operator}: a closed separation is
quotiented away and never revisited.  We now establish four guarantees that the
construction delivers, each isolating one face of the same recursive
refine-condense mechanism.  \emph{Separability} states that quotienting preserves
the ability to separate classes and propagates it up the entire tower;
\emph{stability} states that committed levels are invariant under further
refinement, so composition is interference-free; \emph{bounded capacity} states
that the covering number per level stays bounded under uniform contraction,
independent of instance length; and \emph{scalability} states that inference cost
scales with quotient distance rather than ambient distance.  Throughout,
$N(\epsilon,\mathcal M)$ denotes the $\epsilon$-covering number of a metric space
$\mathcal M$ \cite{kolmogorov1959entropy,shalev2014understanding}, a separator at
level $k$ is a continuous $f_k:\mathcal M_k\to[0,1]$, and its induced classifier is
the threshold rule $\mathbf 1[f_k>\tfrac12]$.

\paragraph{1) Separability}
Classical capacity theory treats separability as a volume problem: in a fixed
$d$-dimensional space only $O(d)$ regions are stably separable
\cite{cover1965geometry}.  The ladder reframes the obstacle as one of
\emph{connectivity} rather than volume: by Urysohn's lemma, separation requires
only a continuous deformation that renders the supports disjoint, and quotienting
supplies exactly such a deformation without raising dimension.

\begin{theorem}[Quotient collapse preserves separability]\label{thm:collapse-sep}
  Let \(\mathcal M\) be a normal space and let \(A,B\subset\mathcal M\) be
  nonempty disjoint closed sets.  By Urysohn's lemma there is a continuous
  \(f:\mathcal M\to[0,1]\) with \(f(A)=\{0\}\) and \(f(B)=\{1\}\).  Let
  \(x\sim_f y \iff f(x)=f(y)\) and let
  \(q:\mathcal M\to\widetilde{\mathcal M}:=\mathcal M/\!\sim_f\) be the quotient
  map.  Then \textup{(1)} \(q(A)\) and \(q(B)\) are distinct points of
  \(\widetilde{\mathcal M}\); \textup{(2)} there is a unique continuous
  \(\bar f:\widetilde{\mathcal M}\to[0,1]\) with \(f=\bar f\circ q\),
  \(\bar f(q(A))=0\), \(\bar f(q(B))=1\); and \textup{(3)} \(q(A),q(B)\) are
  separated by the threshold rule \(\mathbf 1[\bar f>\tfrac12]\).
\end{theorem}

A single collapse suffices for one level; and the construction of Urysohn Ladder guarantees the separation across the hierarchy, The next result shows the property is
hereditary, so a classifier built at the base survives every condensation.

\begin{theorem}[Separation propagates through the ladder]\label{thm:ladder-sep}
  Let \(\mathcal M_0\) be normal with nonempty disjoint closed \(A_0,B_0\), and let
  \(q_k:\mathcal M_k\to\mathcal M_{k+1}\) be the ladder's quotient maps with
  \(A_{k+1}:=q_k(A_k)\), \(B_{k+1}:=q_k(B_k)\).  If the base separator
  \(f_0:\mathcal M_0\to[0,1]\) is compatible with the tower, meaning that at each
  level \(f_k\) is constant on the equivalence classes collapsed by \(q_k\), then
  for every \(k\) there is a continuous \(f_k:\mathcal M_k\to[0,1]\) separating
  \(A_k\) and \(B_k\).
\end{theorem}

\begin{remark}
  Compatibility is the load-bearing hypothesis: it requires each condensation to
  respect the existing separation (collapse only \emph{within} a class, never
  across the boundary).  This is precisely the containment discipline of the stack
  (\Cref{rem:stack-savitch}); a quotient that violated it would correspond to
  popping past a committed level.
\end{remark}

\paragraph{2) Stability}
\Cref{thm:ladder-sep} guarantees a separator survives condensation but not that
condensing a \emph{new} region leaves \emph{previously} committed tokens
undisturbed.  Stability is the guarantee that refinement and commitment occupy
orthogonal parts of the state, so that the two cannot interfere.  Write the state
parameters as a pair $\theta=(\theta_F,\theta_S)\in\Theta_F\times\Theta_S$: a
\emph{flow} block $\theta_F$ carrying the active, plastic refinement, and a
\emph{scaffold} block $\theta_S$ carrying the frozen, committed tokens.

\begin{theorem}[Refinement does not perturb committed tokens]\label{thm:stability}
  Suppose the metric on parameter space is block-diagonal, $g=g_F\oplus g_S$, so
  $\langle(u_F,0),(0,v_S)\rangle_g=0$, and suppose the ladder alternates a
  \emph{flow} update ($\Delta\theta_S=0$) with a \emph{condensation} update
  ($\Delta\theta_F=0$).  Then the cross term vanishes,
  $\langle\Delta\theta^{(F)},\Delta\theta^{(S)}\rangle_g=0$, and any committed
  functional $R$ depending only on $\theta_S$ is invariant under flow updates:
  $R(\theta_F+\Delta\theta_F,\theta_S)=R(\theta)$.
\end{theorem}

\begin{remark}[Stability-Plasticity Tradeoff]
  Given the decomposition, the conclusion is elementary (an orthogonality
  computation).  The content is therefore not in the proof but in the
  \emph{architectural} claim that the ladder maintains such a decomposition: the
  freezing of \Cref{def:operator} is exactly what confines plastic updates to
  $\theta_F$ and renders the committed $\theta_S$ inert.  Under this discipline,
  composition is interference-free \cite{french1999catastrophic}, committed separations are reused without being
  overwritten, which is the formal statement of the structural-decoupling property
  the framework relies on.  The guarantee is only as strong as the decomposition:
  where flow and scaffold are not cleanly separated (overlapping supports,
  \Cref{ssec:limits}), the cross term reappears - plasticity gains at the price of degraded stability \cite{mccloskey_catastrophic_1989}.
\end{remark}

\paragraph{3) Bounded Capacity}
Separability guarantees a boundary exists after collapse; viability further
requires the quotient tower to be \emph{compact}, so that resource demand scales
with intrinsic complexity rather than instance length.  Take the effective
capacity demand of $\mathcal M$ at resolution $\epsilon$ to be its covering number
$N(\epsilon,\mathcal M)$; a system with budget $d$ can represent $\mathcal M$ only
if $N(\epsilon,\mathcal M)\le d$. We formalize this observation with the following lemma.

\begin{lemma}[Linear growth without contraction]\label{lem:linear-growth}
  If $\mathcal M_0$ is isometric to a segment of length $L$ (no contraction), then
  $N(\epsilon,\mathcal M_0)=\Theta(L/\epsilon)$.
\end{lemma}

\begin{definition}[Recursive $\rho$-compressibility]\label{def:rho}
  The tower is \emph{recursively $\rho$-compressible} if there is a uniform
  $\rho>1$ with $N(\epsilon,\mathcal M_{k+1})\le\rho^{-1}N(\epsilon,\mathcal M_k)$
  for all $k$.
\end{definition}

\begin{theorem}[Bounded capacity under recursive contraction]\label{thm:capacity}
  Under \Cref{def:rho}, \(N(\epsilon,\mathcal M_D)\le\rho^{-D}N(\epsilon,\mathcal
  M_0)\) for all \(D\).  Consequently, to meet a budget \(N(\epsilon,\mathcal M_D)\le
  d\), it suffices to take
    $D\ \ge\
    \max\!\left\{0,\,
    \Bigl\lceil \log_\rho\!\bigl(N(\epsilon,\mathcal M_0)/d\bigr)\Bigr\rceil
    \right\}$.
  Hence bounded representational demand is maintained for arbitrarily long
  instances by logarithmic growth in depth whenever the initial covering demand
  grows at most polynomially with instance length.
\end{theorem}

\begin{remark}
  The theorem converts a uniform per-level compression into logarithmic depth; the
  assumption $\rho>1$ is doing the work and encodes that the environment has
  nested, recursively condensable structure \cite{simon1973architecture}.  Where $\rho\to1$ (no genuine
  abstraction is available), depth provides no relief and capacity demand reverts
  to the linear growth of \Cref{lem:linear-growth}.  This is the capacity-side
  counterpart of the depth/space economy of \Cref{rem:stack-savitch}.
\end{remark}

\paragraph{4) Scalability}
Finally, we account for the cost of \emph{using} the ladder.  The contraction that
bounds capacity also bounds the variation of the separator inside each collapsed
cell, which is what makes navigation on the quotient faithful.

\begin{lemma}[Compatibility implies local boundedness]\label{lem:local-bound}
  Let $q:X\to X/\!\sim$ have classes of diameter $\le\varepsilon$, let $f:X\to[0,1]$
  be constant on classes, and let $f$ be $L$-Lipschitz.  Then $f$ descends to a
  unique continuous $\bar f$ on the quotient, and $|f(u)-f(x)|\le L\varepsilon$ for
  $u\in[x]$: the score varies by at most $L\varepsilon$ within any cell.
\end{lemma}


Because each cell is internally near-constant, inference need not traverse the
ambient trajectory: it navigates the quotient graph $G_Q$ whose vertices are
tokens.  Let the cost of planning within any one cell (entering and exiting
through admissible boundary states) be bounded by a constant $\xi$, and let
$d_{G_Q}$ denote shortest-path distance on $G_Q$.

\begin{theorem}[Quotient-distance scaling]\label{thm:scaling}
  For any source and goal \(s,g\) with token classes \(q(s),q(g)\),
    $E_{\mathrm{total}}(s,g)\ \le\
    \xi\cdot d_{G_Q}\!\bigl(q(s),q(g)\bigr)\;+\;E_{\mathrm{hi}}\!\bigl(q(s),q(g)\bigr)
    \;+\;E_{\mathrm{bdry}}$,
  where \(E_{\mathrm{hi}}\) is the cost of finding a path in \(G_Q\) and
  \(E_{\mathrm{bdry}}\) is a boundary-crossing overhead.  If
  \(E_{\mathrm{hi}}=O(d_{G_Q})\) and \(E_{\mathrm{bdry}}=O(1)\), then
    $E_{\mathrm{total}}(s,g)=O\!\bigl(d_{G_Q}(q(s),q(g))+1\bigr)$.
  If, in addition, each quotient hop has cost bounded below by a positive
  constant, then
    $E_{\mathrm{total}}(s,g)=\Theta\!\bigl(d_{G_Q}(q(s),q(g))+1\bigr)$.
  Thus inference cost scales with quotient distance rather than with the ambient
  trajectory length between \(s\) and \(g\).
\end{theorem}

\begin{remark}[The time-side of the Savitch economy]
  Together with \Cref{rem:stack-savitch}, this completes the depth picture of
  Section~\ref{sec:dynamics}.  Bounding the diameter by $\rho^{-D}$ forces
  $D=O(\log L)$; \Cref{thm:scaling} then shows inference traverses $O(\log L)$
  tokens rather than $O(L)$ ambient steps.  Where the stack-depth economy bounded
  \emph{space}, quotient-distance scaling bounds \emph{inference cost}: condensation
  is a geometric memoization that converts repeated recursive verification into
  direct navigation on the folded manifold, the geometric reading of Savitch-style
  recursion \cite{savitch1970relationships,burago2001course}.
\end{remark}

\paragraph{Scope of the guarantees}
The four guarantees are consequences of one mechanism, recursive, compatible
quotienting, and each carries one load-bearing hypothesis: compatibility
(\Cref{thm:ladder-sep}), the flow/scaffold decomposition (\Cref{thm:stability}),
uniform $\rho$-compression (\Cref{thm:capacity}), and bounded per-cell cost
(\Cref{thm:scaling}).  All four share a single failure mode: \emph{overlapping
supports}.  When the regions to be collapsed are not cleanly disjoint,
compatibility fails, the flow/scaffold split leaks, $\rho\to1$, and within-cell
variation is no longer bounded (so separability, stability, capacity, and
scalability degrade in concert).  The guarantees are therefore tight exactly in the
well-nested regime in which the ladder was constructed, and quantifying their
degradation as supports overlap is the central open direction.

\section{Discussion and Conclusion}\label{sec:future}

The~\UM{} formalizes a way of viewing classification as topological indexing
followed by local metric prediction.  Its main point is that a symbolic
simulation can hide a cost that becomes explicit in the metric formulation: the
amount of boundary structure that must be constructed, stored, and reused.
Urysohn Width is intended to expose this hidden cost.  This distinction
clarifies the role of amortization in cognitive computation
\cite{gershman_amortized_2014,ritchie2016deep}.  A difficult classification
boundary may require substantial exploratory work before the appropriate
separator is found.  Once represented as a Urysohn Triple, however, the same
boundary can be reused through library lookup and stack composition.  The model
therefore separates the cost of \emph{constructing} a separator from the cost of
\emph{executing} it after construction.  This separation is natural in learning
systems, where training, consolidation, and inference need not have the same
resource profile.

There are several limitations.  First, Urysohn Width is a geometric measure and
depends on the chosen metric representation.  Different embeddings of the same
abstract classification problem can induce different boundary measures.  Second,
the present paper develops the theory primarily for idealized metric libraries
and exact stack operations.  Approximate libraries, noisy separators, and
data-dependent metric learning require additional stability results.  Third, the
consistency of the boundary-width estimator in \Cref{thm:cut-consistency}
assumes a \(C^{1,1}\) boundary, a density bounded away from zero, and the
connectivity scaling; rougher boundaries, anisotropic sampling, and
finite-sample rates would require quantitative concentration bounds on the
empirical cut functional, which we leave to future work.

The broader implication is that classical computability and metric
classification need not be opposed.  The~\TM{} gives an extensional theory of
what can be computed: it characterizes the input--output functions realizable by
effective procedures \cite{turing1936computable}.  The~\UM{}, by contrast,
gives an intensional account of how metric-topological structure can be
represented, amortized, and reused.  In this sense, the~\UM{} is best viewed as
a refinement of computational description: it preserves the standard boundary of
computability while providing a more structured language for problems whose
difficulty is governed by geometry.

This distinction is especially relevant for continual learning
\cite{kirkpatrick2017overcoming,parisi2019continual,delange2021continual}.  In
a purely parametric system, learning a new task often means modifying the same
global weights that support previous tasks, creating the familiar problem of
interference or catastrophic forgetting.  In the~\UM{} view, a new task is
instead a new or refined separator inserted into a Metric Library.  Once a
frontier has been committed, it can be frozen, reused, or refined without
requiring the entire system to be rewritten.  Continual learning therefore
becomes a problem of controlled frontier refinement: the learner must decide
when a new observation lies inside an existing basin, when it requires local
extension, and when it requires a genuinely new separator.  This turns memory
from a passive store of examples into an active geometry of reusable
distinctions.

The same perspective also clarifies the memory wall in large-scale learning.
Empirical scaling laws show that larger models can absorb more structure when
model size, data, and compute are jointly scaled
\cite{kaplan2020scaling,hoffmann2022training}.  However, the cost of retrieving
and updating that structure increasingly depends on memory movement, context
length, and reuse of previously computed representations
\cite{wulf1995hitting,horowitz2014computing,dao2022flashattention}.  A
Urysohn-style library separates the one-time cost of constructing a frontier
from the repeated cost of querying it.  Once a boundary has been realized, later
inference can proceed by contracted geodesics and quotient-level lookup rather
than by rediscovering the same separation inside a monolithic parameter space.
Thus the relevant resource is not only parameter count or floating-point
operations, but the amount of boundary structure that must be stored, indexed,
and traversed.  Urysohn Width gives a language for this resource: it measures
the frontier mass that must be maintained for reliable reuse.

For this reason, the~\UM{} also suggests a conservative route toward AGI.  It
does not claim that intelligence requires computation beyond the Turing limit.
Rather, it suggests that general intelligence may require a richer internal
organization of computable structure: stable separators for abstraction,
frontier-preserving extensions for generalization, explicit rejection of
out-of-domain extrapolation, and reusable metric contractions for amortized
inference.  This is consistent with views of intelligence that emphasize
generalization, compositionality, model-based structure, and reusable internal
representations \cite{legg2007universal,lake2017building,sutton2019bitter}.  An
AGI system, in this view, is not merely a larger sequence model but a system that
can continually build, protect, compose, and query a growing library of
metric-topological distinctions.  The Turing machine remains the theory of
effective computation; the Urysohn machine refines the theory of effective
representation for agents that must learn continuously, remember selectively,
generalize safely, and act under geometric constraints.

\begin{acks}
This work was partially supported by NSF grants IIS-2401748 and BCS-2401398.
The author used ChatGPT 5.5 and Claude Opus 4.8 for language polishing,
mathematical proofs, and visual-illustration drafting.  All theoretical claims,
conceptual ideas, and final manuscript decisions are the author's responsibility.
\end{acks}

\begin{center}
    \large Appendix
\end{center}

\begin{appendix}

\begin{proof}[Proof of Theorem \ref{thm:realization}]
Let
  $\mathbb D=\{m2^{-q}:m,q\in\mathbb N,\ 0\le m2^{-q}\le 1\}$
be the dyadic rationals in \([0,1]\).  We first construct the dyadic family.
By T1 in Definition \ref{def:dyadic-normality}, choose \(P_0\in\mathcal R\) with
\(A\subset \operatorname{Int}_X P_0\) and \(P_0\cap B=\varnothing\).  We build,
by induction on \(n\), sets \(P_r\in\mathcal R\) for all
\(r\in\mathbb D_n:=\{m2^{-n}:0\le m<2^n\}\) such that
  $r<s \quad\Longrightarrow\quad P_r\Subset P_s,
  \quad\text{and}\quad
  P_r\cap B=\varnothing$.
The claim is true for \(n=0\).  Suppose it holds at level \(n\).  For every
adjacent pair \(r<s\) in \(\mathbb D_n\), T2 gives a set
\(P_{(r+s)/2}\in\mathcal R\) with
  $P_r\Subset P_{(r+s)/2}\Subset P_s$.
For the terminal dyadic point \(1-2^{-(n+1)}\), T3 applied to
\(P_{1-2^{-n}}\) gives a set \(Q\in\mathcal R\) with
  $P_{1-2^{-n}}\Subset Q,
  \quad
  Q\cap B=\varnothing$.
Set \(P_{1-2^{-(n+1)}}=Q\).  This completes the induction and yields a family
\(\{P_r\}_{r\in\mathbb D\cap[0,1)}\) satisfying
  $A\subset \operatorname{Int}_X P_0,\quad
  P_r\cap B=\varnothing,\quad
  r<s\Rightarrow P_r\Subset P_s$.
In particular, \(P_r\subset \operatorname{Int}_X P_s\) whenever \(r<s\).
Set
  $U_r:=\operatorname{Int}_X P_r$.
Define
  $f(x)=\inf\{r\in\mathbb D\cap[0,1):x\in U_r\}$,
with the convention that the infimum of the empty set is \(1\).  Since the index
set lies in \([0,1)\), this gives a well-defined map \(f:X\to[0,1]\).
If \(x\in A\), then \(x\in U_0\), so the defining set contains \(0\), and hence
\(f(x)=0\).  If \(x\in B\), then \(x\notin P_r\), and therefore \(x\notin U_r\),
for every \(r<1\).  The defining set is empty and \(f(x)=1\).  Thus
\(f|_A=0\) and \(f|_B=1\).

We prove continuity by computing preimages of a subbasis of the order topology
on \([0,1]\).  For \(0<a\le1\),
  $f^{-1}([0,a))
  =
  \bigcup_{\substack{r\in\mathbb D\cap[0,1)\\ r<a}} U_r$.
The inclusion ``\(\supseteq\)'' follows from \(f(x)\le r<a\).  Conversely, if
\(f(x)<a\), then by the definition of infimum there exists dyadic \(r<a\) with
\(x\in U_r\).  Hence this preimage is open.  For \(0\le a<1\),
  $f^{-1}((a,1])
  =
  \bigcup_{\substack{r\in\mathbb D\cap[0,1)\\ r>a}} (X\setminus P_r)$.
If \(x\notin P_r\) for some \(r>a\), then for every dyadic \(q\le r\) one has
\(U_q\subset P_q\subset P_r\), so \(x\notin U_q\) and \(f(x)\ge r>a\).
Conversely, if \(f(x)>a\), choose dyadic \(r,t\) with
  $a<r<t<f(x)$.
If \(x\in P_r\), then \(P_r\subset U_t\), hence \(x\in U_t\) and
\(f(x)\le t<f(x)\), a contradiction.  Thus \(x\notin P_r\).  Since each
\(P_r\) is closed, the displayed union is open.  Therefore \(f\) is continuous.

It remains to prove the chain identities.  By chain-level frontier realization,
each \(P_r\) has an associated chain \(c_{P_r}\) whose boundary realizes the
frontier.  For any finite set of indices, we regard the chains on a common
subdivision of \(K\), which is allowed because finite polyhedral subdivisions
admit a common refinement and simplicial chains push forward to such a
refinement.  Define
  $\Gamma_r:=\partial c_{P_r}$.
Since \(\partial^2=0\),
  $\partial\Gamma_r=\partial^2 c_{P_r}=0$,
so every dyadic frontier is a cycle.  For \(r<s\), define
  $s_{r,s}:=c_{P_s}-c_{P_r}$.
By linearity of \(\partial\),
  $\partial s_{r,s}
  =
  \partial c_{P_s}-\partial c_{P_r}
  =
  \Gamma_s-\Gamma_r$.
Applying \(\partial\) once more gives
  $\partial(\Gamma_s-\Gamma_r)=\partial^2 s_{r,s}=0$.
Thus the separator is accompanied by the asserted chain-level frontier calculus.
\end{proof}

\begin{proof}[Proof of Theorem \ref{thm:cut-consistency}]
Let \(X_1,\ldots,X_n\) be independent samples from density \(p\).  The diagonal
terms vanish because \(f(X_i)=f(X_i)\), so replacing \(n^2\) by \(n(n-1)\)
changes the estimator by a factor tending to \(1\).  Define the symmetric
triangular-array kernel
  $H_h(x,y)
  :=
  \frac{1}{\sigma_\eta h^{d+1}}\,
  \eta\!\left(\frac{\|x-y\|}{h}\right)\mathbf 1[f(x)\neq f(y)]$.
Then \(\widehat{\Width}_{\partial,h}\) is asymptotically the corresponding
second-order \(U\)-statistic.

We first compute its expectation.  With the change of variables \(y=x+hz\),
  $\mathbb E H_h(X,Y)
  =
  \frac{1}{\sigma_\eta h}
  \int_{\Sigma}\int_{\R^d}
  \eta(|z|)\mathbf 1[f(x)\neq f(x+hz)]p(x)p(x+hz)\,dz\,dx$.
Because the two labels differ only when the segment from \(x\) to \(x+hz\)
crosses \(\partial\tau\), the integrand is supported in an \(O(h|z|)\)-tube
around \(\partial\tau\) \cite{garcia-trillos-slepcev-2016}.  Since \(\partial\tau\) is \(C^{1,1}\), the tubular
neighborhood theorem gives coordinates \(x=u+t\nu(u)\), where
\(u\in\partial\tau\), \(|t|\le Ch\), and \(\nu(u)\) is a measurable unit normal.
In these coordinates, for fixed \(z\),
  $\frac1h\,\mathbf 1[f(u+t\nu(u))\neq f(u+t\nu(u)+hz)]
  \longrightarrow
  |\langle z,\nu(u)\rangle|$
in \(L^1(dt\,d\Hdone(u))\).  The boundedness of \(p\), the compact support of
\(\eta\), and dominated convergence therefore yield
  $\mathbb E\,\widehat{\Width}_{\partial,h}
  \longrightarrow
  \frac1{\sigma_\eta}
  \int_{\partial\tau}p(u)^2
  \int_{\R^d}\eta(|z|)|\langle z,\nu(u)\rangle|\,dz\,d\Hdone(u)$.
The inner integral is \(\sigma_\eta\) by radial symmetry, so the limit is
  $\int_{\partial\tau}p(u)^2\,d\Hdone(u)$.

It remains to control fluctuations.  Since \(\eta\) is bounded and compactly
supported,
  $|H_h(x,y)|\le C h^{-(d+1)}$.
Hoeffding's inequality for bounded \(U\)-statistics gives \cite{hoeffding1963probability}, for each fixed
\(t>0\),
  $\Pr\!\left(
  \left|\widehat{\Width}_{\partial,h_n}
  -\mathbb E\widehat{\Width}_{\partial,h_n}\right|>t
  \right)
  \le
  2\exp\{-c\,n h_n^{2d+2}t^2\}$.
Assumption \textup{(A4)} makes the right-hand side summable in \(n\).  The
Borel-Cantelli lemma implies almost-sure convergence of the estimator to its
expectation, and the expectation limit above proves the first claim.
For the density-corrected estimator, \(p_{\min}>0\) and uniform consistency of
\(\hat p\) imply
  $\sup_x\left|\frac{p(x)}{\hat p(x)}-1\right|\to0
  \quad\text{almost surely}$.
Thus the same argument applies with the factor \(p(x)p(y)\) cancelled by
\(\hat p(x)\hat p(y)\), giving the unweighted limit
\(\Hdone(\partial\tau)=\Width_{\partial}(\tau)\).
\end{proof}

\begin{proof}[Proof of Proposition \ref{prop:cs-spectrum}]
For the unnormalized graph Laplacian,
  $u^\top L_{\rm un}u
  =
  \frac12\sum_{i,j}W_{ij}(u_i-u_j)^2$.
Hence \(u\in\ker L_{\rm un}\) if and only if \(u_i=u_j\) whenever \(i\) and
\(j\) are connected by an edge of positive weight.  Equivalently, \(u\) is
constant on each connected component of \(G_h\).  The indicator functions of the
connected components form a basis for the kernel, so
  $\dim\ker L_{\rm un}=b_0(G_h)$.
For each connected component \(H\), the normalized Laplacian of \(H\) has a
simple zero eigenvalue and first nonzero eigenvalue \(\lambda_2(H)\).  The
standard two-sided Cheeger inequality for weighted graphs gives
  $\frac{\lambda_2(H)}{2}\le \Phi(H)\le \sqrt{2\lambda_2(H)}$.
This proves the conductance statement.  Finally, if a perturbation changes the
normalized Laplacian by \(E\) with \(\|E\|_{\rm op}\le\delta\), and if the gap
between the zero eigenspace and the rest of the spectrum is \(\gamma>0\), the
Davis--Kahan sin-theta theorem bounds the difference between the corresponding
spectral projectors by \(O(\delta/\gamma)\).  This proves the stated stability
claim.
\end{proof}

\begin{proof}[Proof of Theorem \ref{thm:amortized}]
By the boundary-coverage hypothesis,
  $\partial\tau\subseteq \bigcup_{i=1}^k U_i$.
Using countable subadditivity of \((d-1)\)-dimensional Hausdorff measure,
  $\Width_{\partial}(\tau)
  =
  \Hdone(\partial\tau)
  \le
  \sum_{i=1}^k \Hdone(\partial\tau\cap U_i)$.
The simple-support hypothesis gives
  $\Hdone(\partial\tau\cap U_i)\le C\eps$
for each \(i\).  Therefore
  $\Width_{\partial}(\tau)\le k C\eps$,
which rearranges to
  $k\ge \frac{\Width_{\partial}(\tau)}{C\eps}$.
\end{proof}

\begin{proof}[Proof of Theorem \ref{thm:geodesic}]
Let \(W_i\) denote the amount of boundary mass traversed at level \(i\).  By
hypothesis, the within-class contraction at that level multiplies the relevant
path length by at most \(\lambda_i\).  Thus the level-\(i\) contribution to the
contracted path is at most \(\lambda_i W_i\).  Summing over levels gives
  $\ell_{\rm contracted}\le \sum_{i=1}^d \lambda_i W_i$.
If \(W_i\le \Width_U(\Lib)/d\) for all \(i\), then
  $\ell_{\rm contracted}
  \le
  \sum_{i=1}^d \lambda_i\,\frac{\Width_U(\Lib)}{d}$.
For uniform contraction \(\lambda_i=\lambda<1\), this becomes
  $\ell_{\rm contracted}
  \le
  \lambda\,\Width_U(\Lib)
  <
  \Width_U(\Lib)$.
\end{proof}

\begin{proof}[Proof of Proposition \ref{prop:edt-convergence}]
At the initial scale, all labels lie in \([0,1]\), so the largest unresolved
label gap has width at most \(1\).  By assumption, one full refinement round
subdivides every active gap by a factor at least \(\rho\).  Therefore, after
\(k\) full rounds, every active gap has width at most \(\rho^{-k}\).  Choosing
  $k\ge \left\lceil \log_\rho(1/\varepsilon)\right\rceil$
gives \(\rho^{-k}\le\varepsilon\).  At this point no gap remains above the
resolution threshold \(\varepsilon\), so no refinement is triggered merely by
lack of an intermediate dyadic label.  If individual events refine only one gap,
then the same estimate applies after each active gap has been refined once per
round; hence the logarithmic bound counts complete rounds rather than individual
insertions.
\end{proof}

\begin{proof}[Proof of Corollary \ref{cor:optimal-radix}]
Since \(\ln(1/\varepsilon)\) is independent of \(\rho\), minimizing
  $\mathcal C(\rho)=\rho\,\frac{\ln(1/\varepsilon)}{\ln\rho}$
over \(\rho>1\) is equivalent to minimizing \(g(\rho)=\rho/\ln\rho\).  A direct
derivative calculation gives
  $g'(\rho)=\frac{\ln\rho-1}{(\ln\rho)^2}$.
Thus \(g\) decreases on \((1,e)\), increases on \((e,\infty)\), and has its
unique global minimum at \(\rho=e\).  The corresponding depth is
  $\log_e(1/\varepsilon)=\ln(1/\varepsilon)$,
and the continuous-radix cost is \(e\ln(1/\varepsilon)\).  If \(\rho\) is
restricted to integers \(>1\), then \(g(3)<g(2)\), and since \(g\) is increasing
for \(\rho\ge e\), the integer optimum is \(\rho=3\).
\end{proof}

\begin{proof}[Proof of Theorem \ref{thm:collapse-sep}]
Because \(f\) is constant on each equivalence class of \(\sim_f\), the formula
  $\bar f([x]) := f(x)$
defines a well-defined map \(\bar f:\widetilde{\mathcal M}\to[0,1]\).  Since
\(q\) is the quotient map and \(f=\bar f\circ q\) is continuous, the universal
property of the quotient topology implies that \(\bar f\) is continuous.  The
same formula also proves uniqueness.
If \(a,a'\in A\), then \(f(a)=f(a')=0\), so \(q(a)=q(a')\); hence \(q(A)\) is a
single point.  Similarly, \(q(B)\) is a single point.  These points are distinct:
if \(q(a)=q(b)\) for \(a\in A\), \(b\in B\), then \(f(a)=f(b)\), contradicting
\(0= f(a)\neq f(b)=1\).  Finally,
  $\bar f(q(A))=0,\qquad \bar f(q(B))=1$,
so the threshold rule \(\mathbf 1[\bar f>\tfrac12]\) assigns different labels to
the two quotient points.
\end{proof}

\begin{proof}[Proof of Theorem \ref{thm:ladder-sep}]
We argue by induction on \(k\).  The case \(k=0\) holds by the assumed base
separator \(f_0\).  Suppose \(f_k:\mathcal M_k\to[0,1]\) is continuous and
separates \(A_k\) from \(B_k\), and suppose \(f_k\) is constant on the
equivalence classes collapsed by \(q_k\).  Define
  $f_{k+1}(q_k(x)):=f_k(x)$.
Compatibility makes this definition independent of the representative \(x\).
Since \(q_k\) is a quotient map and \(f_k=f_{k+1}\circ q_k\), the universal
property of the quotient topology implies that \(f_{k+1}\) is continuous.
Moreover,
  $f_{k+1}(A_{k+1})=f_{k+1}(q_k(A_k))=f_k(A_k)=\{0\}$,
and similarly \(f_{k+1}(B_{k+1})=\{1\}\).  Thus \(f_{k+1}\) separates
\(A_{k+1}\) and \(B_{k+1}\).  In particular, the two sets remain disjoint, since
identifying a point of \(A_k\) with a point of \(B_k\) would force
\(0=f_k(a)=f_k(b)=1\).  The induction proves the claim for all \(k\).
\end{proof}

\begin{proof}[Proof of Theorem \ref{thm:stability}]
Write a flow update as \(\Delta\theta^{(F)}=(\Delta\theta_F,0)\) and a
condensation update as \(\Delta\theta^{(S)}=(0,\Delta\theta_S)\).  Since the
metric is block diagonal,
  $\langle\Delta\theta^{(F)},\Delta\theta^{(S)}\rangle_g
  =
  \langle \Delta\theta_F,0\rangle_{g_F}
  +
  \langle 0,\Delta\theta_S\rangle_{g_S}
  =
  0 $.
If \(R\) depends only on \(\theta_S\), then for any flow update,
  $R(\theta_F+\Delta\theta_F,\theta_S)=R(\theta_S)=R(\theta_F,\theta_S)$.
Thus committed functionals are invariant under flow updates.
\end{proof}

\begin{proof}[Proof of Lemma \ref{lem:linear-growth}]
Let \(\mathcal M_0\) be isometric to an interval of length \(L\).  An
\(\epsilon\)-ball in this interval covers length at most \(2\epsilon\), so any
\(\epsilon\)-cover needs at least \(L/(2\epsilon)\) balls.  Conversely, placing
centers at spacing \(\epsilon\) gives an \(\epsilon\)-cover with at most
\(\lceil L/\epsilon\rceil+1\) balls.  Therefore
  $\frac{L}{2\epsilon}\le N(\epsilon,\mathcal M_0)
  \le \left\lceil\frac{L}{\epsilon}\right\rceil+1$,
which is \(N(\epsilon,\mathcal M_0)=\Theta(L/\epsilon)\).
\end{proof}

\begin{proof}[Proof of Theorem \ref{thm:capacity}]
We prove the first claim by induction on \(D\).  For \(D=0\) it is an equality.
If it holds for \(D\), then recursive \(\rho\)-compressibility gives
  $N(\epsilon,\mathcal M_{D+1})
  \le
  \rho^{-1}N(\epsilon,\mathcal M_D)
  \le
  \rho^{-(D+1)}N(\epsilon,\mathcal M_0)$.
This proves
  $N(\epsilon,\mathcal M_D)\le \rho^{-D}N(\epsilon,\mathcal M_0)$
for all \(D\).  To enforce \(N(\epsilon,\mathcal M_D)\le d\), it is enough that
  $\rho^{-D}N(\epsilon,\mathcal M_0)\le d$,
or equivalently
  $D\ge \log_\rho\!\bigl(N(\epsilon,\mathcal M_0)/d\bigr)$.
If \(N(\epsilon,\mathcal M_0)\le d\), depth \(D=0\) already suffices; otherwise
rounding up gives the displayed maximum-with-zero bound.
\end{proof}

\begin{proof}[Proof of Lemma \ref{lem:local-bound}]
Since \(f\) is constant on equivalence classes of \(q\), the rule
  $\bar f(q(x)):=f(x)$
is well defined.  The identity \(f=\bar f\circ q\), together with the fact that
\(q\) is a quotient map and \(f\) is continuous because it is Lipschitz, implies
that \(\bar f\) is continuous by the universal property of the quotient
topology.  Uniqueness follows because \(q\) is surjective.
If \(u\in[x]\), then the class diameter assumption gives \(d(u,x)\le
\varepsilon\).  Since \(f\) is \(L\)-Lipschitz,
  $|f(u)-f(x)|\le L\,d(u,x)\le L\varepsilon$.
In fact, because \(f\) is constant on classes, the left-hand side is zero; the
\(L\varepsilon\) bound is the robust form used when cells are only
approximately collapsed.
\end{proof}

\begin{proof}[Proof of Theorem \ref{thm:scaling}]
Choose a shortest path in \(G_Q\) from \(q(s)\) to \(q(g)\).  It has
\(d_{G_Q}(q(s),q(g))\) quotient hops.  By the hypothesis defining \(\xi\), each
within-cell traversal needed to enter, cross, or exit a cell costs at most
\(\xi\).  Therefore the total within-cell cost is at most
  $\xi\,d_{G_Q}(q(s),q(g))$.
Adding the high-level planning cost and the boundary-crossing overhead gives the
displayed upper bound.  If \(E_{\mathrm{hi}}=O(d_{G_Q})\) and
\(E_{\mathrm{bdry}}=O(1)\), this upper bound is \(O(d_{G_Q}+1)\).  If each
quotient hop also has cost at least a fixed constant \(c>0\), then any path with
\(d_{G_Q}\) hops costs at least \(c\,d_{G_Q}\), giving the matching lower bound
and hence the stated \(\Theta(d_{G_Q}+1)\) scaling.
\end{proof}

\end{appendix}

\bibliographystyle{ACM-Reference-Format}
\bibliography{ref,references}

\end{document}